\documentclass[11pt]{article}
\usepackage{acl}
\usepackage{times}
\usepackage{latexsym}
\usepackage[T1]{fontenc}
\usepackage[utf8]{inputenc}
\usepackage{microtype}
\usepackage{inconsolata}
\usepackage{graphicx}
\usepackage{booktabs}
\usepackage{amsmath}
\usepackage{amssymb}
\usepackage{subcaption}
\usepackage{placeins}
\usepackage{float}
\usepackage{enumitem}
\setlist{noitemsep,topsep=2pt}

\title{Sensitivity, Causality, and Repair Dissociate:\\
A Layer-Wise Analysis of Perturbation Robustness and Its Scaling}

\author{
  Nathan Labiosa, David Buff, Ena Nayak, Erica Donno \\
  University of Southern California \\
  \texttt{\{labiosa, dbuff, enayak, donno\}@usc.edu}
}

\begin{document}
\maketitle

\begin{abstract}
When a language model fails on surface-perturbed input (typos, OCR
noise, homophones), ``which layer is responsible'' has three natural
operationalizations: where representations diverge most
(\emph{sensitivity}), where restoring clean activations recovers the
prediction (\emph{causality}), and where a small adapter can repair
the damage (\emph{compensatory capacity})---and we show these three
layer maps dissociate. Across a five-model panel we identify two propagation
regimes---\emph{spike-and-suppress} (Phi-3.5, Gemma-2-9B) and
\emph{late-accumulation} (Llama-3, Mistral, Qwen2.5-7B)---and on the
two models meeting an $80\%$ identity-patch gate, sensitivity and
causality are \emph{anti-correlated} ($\rho = -0.72$ to $-0.88$).
Within-family scaling on Qwen2.5 ($1.5$B$\to$$14$B) shows the
late-accumulation signature strengthening monotonically with scale,
corroborated on a second family. We propose \emph{cascade disruption}
as the mechanism behind the dissociation: adapters placed at causally
implicated early layers break intact downstream computation, making
diagnostic-flagged sites the \emph{worst} adapter placements. A
fixed-harness layer sweep across four models ($3.8$--$8$B) confirms
the core prediction on chain-of-thought GSM8K---the flagged sites are
the most damaging adapter windows on every adjudicable model---and is
sign-consistent but strongly attenuated on a multiple-choice control,
consistent with damage that compounds with generation length. The
sweep yields practical guidance: a training-free LRD pre-screen and a
default-deepest placement rule, though absolute gains over no-adapter
baselines remain small. Finally, apparent gains from a
representation-stability loss reverse under an adequate generation
budget---truncated chain-of-thought had been scored as empty---a
methodological warning for any intervention evaluated on
chain-of-thought tasks.
\end{abstract}

\section{Introduction}
\label{sec:intro}

A language model that solves a math problem will often fail on the
same problem when the input contains a few typos. Surface
perturbations---typos, OCR noise, speech-like substitutions,
homophones---are among the most common corruptions models encounter in
deployment
\cite{pruthi-etal-2019-combating,rijhwani-etal-2020-ocr,liu2023expandingscopeadaptingenglish},
and when they cause a failure, a natural question is \emph{which
layers are responsible}. Mechanistic analysis of transformer internals
\cite{elhage2021mathematical} makes the question tractable, but
``responsible'' hides an ambiguity: it could mean the layers the
perturbation moves most, the layers where restoring the clean
representation recovers the correct output, or the layers where a
lightweight adapter can compensate for the damage.

These three notions feel like they should agree: where the damage is
largest, undoing it ought to help most, and a repair ought to be best
spent. We measure all three per
layer---\textbf{sensitivity}, via Layer-Wise Representation Divergence
(LRD), a cosine divergence between clean and perturbed hidden states
that builds on prior analyses of perturbation propagation
\cite{moradi2021evaluatingrobustnessneurallanguage,shi-huang-2020-robustness};
\textbf{causality}, via activation patching
\cite{meng2023locatingeditingfactualassociations}; and
\textbf{compensatory capacity}, via per-window LoRA adapters---across
a five-model panel. The maps do not agree
(Figure~\ref{fig:threemap_main}). They peak at different layers, and
on the models with the cleanest patching signal, sensitivity and
causality are systematically \emph{anti-correlated}.

The disagreement matters in practice. Interpretability diagnostics
such as divergence probes and activation patching are increasingly
used to decide \emph{where to intervene}: where to place an adapter,
where to apply representation engineering. Our results show that, for
perturbation repair, these diagnostics point at the wrong layers. We
propose a mechanism, \emph{cascade disruption}: an adapter trained at
an early layer perturbs every layer downstream of it, breaking
clean-input computation that was otherwise intact, so repair placed
where the diagnostics point does net harm. A layer sweep under a fixed
evaluation harness confirms this prediction on chain-of-thought
GSM8K; on short-form MMLU the effect is sign-consistent but strongly
attenuated. The flip side is actionable: a training-free LRD
pre-screen rules out early- and mid-layer placement before any
training, and the deepest window is consistently least regressive.
The road to that sweep also produced a methodological
warning: an intervention we ourselves once believed in---a
representation-stability loss with apparently large gains---turned out
to be an artifact of a generation budget too short for
chain-of-thought.

Beyond the dissociation itself, the layer-wise view exposes structure.
Models fall into two propagation regimes---\emph{spike-and-suppress},
where early divergence is absorbed by mid-depth, and
\emph{late-accumulation}, where divergence compounds toward the
output---and within a model family the late-accumulation signature
strengthens monotonically with scale.

Our contributions are:

\begin{enumerate}

\item \textbf{Dissociation.} The three natural per-layer notions of
``responsible for a perturbation failure''---sensitivity, causality,
and compensatory capacity---produce distinct and, in key cases,
anti-correlated layer maps. Evidence: both models passing the $80\%$
identity-patch gate show strong sensitivity--causality
anti-correlation (\S\ref{sec:threemap}); convergent strands in
\S\ref{sec:evidence}.

\item \textbf{Regimes and scaling.} Models split into two propagation
regimes, and the late-accumulation signature strengthens monotonically
with scale within a family. Evidence: a three-point Qwen2.5 scan,
monotone on every perturbation type, corroborated by a two-point Llama
scan (\S\ref{sec:scaling}).

\item \textbf{Mechanism, and a validated placement anti-pattern.}
Cascade disruption---adapters at causally implicated early layers
damage clean-input computation downstream---explains the dissociation
and yields placement guidance: do \emph{not} place adapters where
divergence or patching point; default instead to the deepest window.
Evidence: downstream disruption profiles
(\S\ref{sec:disruption}) and a fixed-harness layer sweep in which
diagnostic-flagged sites are the most damaging windows
(\S\ref{sec:sweep}), width-robust and worst of all when adaptation
spans every layer.

\item \textbf{An evaluation protocol for chain-of-thought
interventions (negative result).} Apparent intervention gains under a
short-generation harness reverse under an adequate generation budget;
the remedy is a generation-length-matched null.
Evidence: the same checkpoint flips from $+7.3$~pp to $-3.5$~pp when
the truncation artifact is fixed (\S\ref{sec:negative}).

\end{enumerate}

\section{Related Work}
\label{sec:related}

\paragraph{Robustness to surface perturbations.}
Character-level and lexical perturbations degrade NLP systems across tasks
\cite{wang-etal-2022-measure, belinkov2018syntheticnaturalnoisebreak,
eger-etal-2019-text, li-etal-2020-bert-attack, dong-smith-2018-multi,
LI2024104598}. Rather than constructing worst-case inputs, we ask where
in the network these failures localize.

\paragraph{Representational similarity.}
Probing \cite{belinkov2021probingclassifierspromisesshortcomings}, CKA
\cite{pmlr-v97-kornblith19a}, SVCCA \cite{NIPS2017_dc6a7e65}, and
model stitching \cite{bansal2021revisitingmodelstitchingcompare} measure what
models encode at each layer. LRD instead tracks
\emph{perturbation-induced} change at every layer, giving a
propagation trajectory rather than a snapshot.

\paragraph{Layer-wise function in transformers.}
Transformer layers are functionally heterogeneous; early layers handle surface
structure and later layers encode more abstract semantic information
\cite{tenney-etal-2019-bert, jawahar-etal-2019-bert, geva-etal-2022-transformer,
geva-etal-2023-dissecting, dar2023analyzingtransformersembeddingspace}. The
three-map dissociation adds a perturbation-robustness dimension to
this heterogeneity.

\paragraph{Activation patching and causal tracing.}
Replacing activations at specific layers with clean-run values is standard for
causal localization \cite{NEURIPS2020_92650b2e,meng2023locatingeditingfactualassociations,
wang2022interpretabilitywildcircuitindirect, hanna2023doesgpt2computegreaterthan,
conmy2023automatedcircuitdiscoverymechanistic}. We show that causal
localization does not transfer to repair: the layers patching
implicates are the most damaging adapter sites in our sweep.

\section{Method}
\label{sec:method}

\subsection{Layer-Wise Representation Divergence (LRD)}

For mean-pooled hidden states $\bar{\mathbf{h}}_{\text{clean}}^{(L)}$
and $\bar{\mathbf{h}}_{\text{noisy}}^{(L)}$ at layer $L$ on a
(clean, perturbed) input pair,
\begin{equation}
\text{LRD}(L) = 1 - \cos\!\bigl(\bar{\mathbf{h}}_{\text{clean}}^{(L)},\,
\bar{\mathbf{h}}_{\text{noisy}}^{(L)}\bigr).
\label{eq:lrd}
\end{equation}
Two summary statistics---\emph{slope} (linear trend across layers) and
\emph{recovery rate}
$\bigl(\text{LRD}(0)-\text{LRD}(L_{\max})\bigr)/\text{LRD}(0)$---distinguish
spike-and-suppress (high recovery, non-positive slope) from
late-accumulation (near-zero recovery, positive slope);
\S\ref{sec:claim1} applies this classification to the panel.

\subsection{Perturbation Types}

We use six perturbation types, five applied at character-level
corruption rates---typos (5\%), OCR (5\%), whitespace (10\%), case
swap (10\%), speech-like (10\%)---and homophones applied at lexical
(per-word) rates (20\%/40\%/50\%). Unless noted,
``Homophones'' cells and the layer-sweep mean use the $20\%$ severity.

\subsection{Activation Patching}
At each layer $L$ we substitute the clean hidden state into the noisy
forward pass and record the fraction of originally-failed examples
(clean-correct, perturbed-incorrect) that now succeed. An identity
(all-layer) patch gives the ceiling; a random-noise patch the floor.
Layers with substantial
single-layer recovery form the \textbf{causal window}. Because a low
identity ceiling means the patching procedure itself is unreliable on
that model, we gate results by it: identity recovery $\geq 80\%$
marks a model as \emph{confirmed}, otherwise \emph{exploratory}.

\subsection{Adapter Setup and Three-Map Signals}
\label{sec:lora_method}

We attach LoRA adapters \cite{hu2021loralowrankadaptationlarge}
($r{=}4$, $\alpha{=}8$), whose per-layer placement flexibility lets us
measure compensatory capacity per layer, to attention
\texttt{q\_proj}/\texttt{v\_proj} (Phi-3.5 has fused attention, so we
use \texttt{qkv\_proj}/\texttt{o\_proj} equivalently) with the base
model frozen, training on perturbed GSM8K with clean supervision
targets and cross-entropy on answer positions. The principal sweep is CE-only; a
representation-stability variant adding
$L_{\text{stab}} = 1 - \cos(\bar{\mathbf{h}}^{\text{LoRA}},
\bar{\mathbf{h}}^{\text{frozen}})$ appears only for the ablation in
\S\ref{sec:negative}. The three-map analysis compares three normalized
per-layer signals: \emph{sensitivity} ($\text{LRD}(L)$), \emph{causality}
(patching recovery at $L$), and \emph{compensatory capacity}
(perturbed-accuracy gain from a 5-layer LoRA window centered at $L$).
Pairwise Spearman $\rho$ is computed across layers per model; per-layer
Cohen's $d$ \cite{cohen1988statistical} at $L$ is the standardized mean
difference in $\text{LRD}(L)$ between failing and succeeding examples.

\subsection{Cascade Disruption and Statistical Framework}
\label{sec:stats_method}

Cascade disruption (\S\ref{sec:disruption}) is the downstream LRD
between frozen and adapted models on clean inputs. Two post-hoc
intrinsic metrics probe whether frozen-model statistics predict
adapter placement: \textbf{C3}, the entropy-based stable rank of the
layer's \texttt{q\_proj}/\texttt{v\_proj} weights, and \textbf{C4},
the Frobenius norm of the teacher-forced cross-entropy gradient on the
same weights (definitions, implementation details, and two further
metrics, C1/C2, in Appendix~\ref{sec:intrinsic}). C3/C4 correlated
with LoRA effectiveness on the original three models but failed
prospective validation (\S\ref{sec:threemap}); we report them as
descriptive only. Because layer-wise signals
autocorrelate (partial-autocorrelation decay to the $95\%$ white-noise
band by lags~$5$--$8$), we use the \textbf{moving-block bootstrap}
\cite{kunsch1989jackknife} (block size~$5$, $10$k resamples) for
correlation CIs and $p$-values; post-decorrelation effective $N$ is
typically $3$--$5$.

\section{Diagnostic Results}
\label{sec:diagnostics}

\paragraph{Setup.} Diagnostic runs use $n{=}500$ per condition
($n{=}200$ for the Llama-3 and Gemma-2-9B recovery profiles;
Table~\ref{tab:supp_recovery}) on
Phi-3.5-mini-instruct ($3.8$B; \citealp{abdin2024phi3technicalreporthighly}),
Mistral-7B-Instruct-v0.3 \cite{jiang2023mistral7b},
Llama-3-8B-Instruct \cite{grattafiori2024llama3herdmodels},
Gemma-2-9B base \cite{gemmateam2024gemma2improvingopen}, and
Qwen2.5-7B-Instruct \cite{qwen2025qwen25technicalreport} over GSM8K
\cite{cobbe2021trainingverifierssolvemath}, MMLU
\cite{hendrycks2021measuringmassivemultitasklanguage}, and BBH
\cite{suzgun2022challengingbigbenchtaskschainofthought}. Clean GSM8K
accuracies are in Table~\ref{tab:regime_summary}. A single fixed
evaluation harness with sufficient
\texttt{max\_new\_tokens} is used throughout (\S\ref{sec:negative});
full hyperparameters are in Appendix~\ref{sec:repro}
(Table~\ref{tab:supp_repro}).

\subsection{Two Propagation Regimes}
\label{sec:claim1}

LRD trajectories across the panel take exactly two shapes. In
\emph{spike-and-suppress} models, perturbation-induced divergence is
large at the embedding and early layers and is then absorbed (a fifth
to three fifths gone by the final layer). In
\emph{late-accumulation} models, early divergence is small but
compounds with depth and never recovers.
Table~\ref{tab:regime_summary} classifies the five models by recovery
rate and slope (\S\ref{sec:method}). LRD heatmaps for
all five models are in Appendix~\ref{sec:lrd_extra}
(Figures~\ref{fig:heatmap}--\ref{fig:supp_lrd}).

\begin{table}[t]
\centering
\small
\begin{tabular}{llccc}
\toprule
Model & Regime & Clean & Recovery & Slope \\
\midrule
Phi-3.5 & S\&S & 85.4\% & 34--60\% & ${\leq}0$ \\
Gemma-2-9B & S\&S & 64.8\% & 21--51\% & ${\leq}0$ \\
Llama-3 & Late & 78.4\% & 0--5\%$^{\S}$ & $+$ \\
Mistral & Late & 59.6\% & 0--5\%$^{\S}$ & $+$ \\
Qwen2.5-7B & Late & 89.0\% & 0--5\%$^{\S}$ & $+$ \\
\bottomrule
\end{tabular}
\caption{Five-model regime classification. S\&S = spike-and-suppress.
Recovery is the fraction of layer-0 divergence eliminated by the final
layer (range across the six perturbation types); slope is the sign of
the linear trend across layers.
$\S$~Homophones ($20\%$ severity) are the one exception: recovery
${\approx}35$--$40\%$ on all three late-accumulation models
(Table~\ref{tab:supp_recovery}); every other late-accumulation cell is
near-zero.}
\label{tab:regime_summary}
\end{table}

Per-perturbation recovery rates confirm the classification
(Appendix~\ref{sec:per_pert}, Table~\ref{tab:supp_recovery}). The
regime labels raise an immediate confound---within this panel, regime
tracks model family---which the within-family scaling scans below are
designed to weaken.

\subsubsection{Within-Family Scaling of the Late-Accumulation Regime}
\label{sec:scaling}

Holding family fixed and varying only scale de-confounds regime from
architecture. The primary scan is Qwen2.5 at $1.5$/$7$/$14$B (a
$9{\times}$ range, all with $\geq 28$ layers); a two-point Llama
corroboration ($1$/$8$B) gives a second family with the same direction
modulo a narrow-bin caveat at the $17$-layer $1$B point
(Figure~\ref{fig:qwen_scaling}; per-perturbation grid in
Appendix~\ref{sec:scaling_extra}, Table~\ref{tab:scaling}).

\begin{figure*}[t]
\centering
\includegraphics[width=0.55\textwidth]{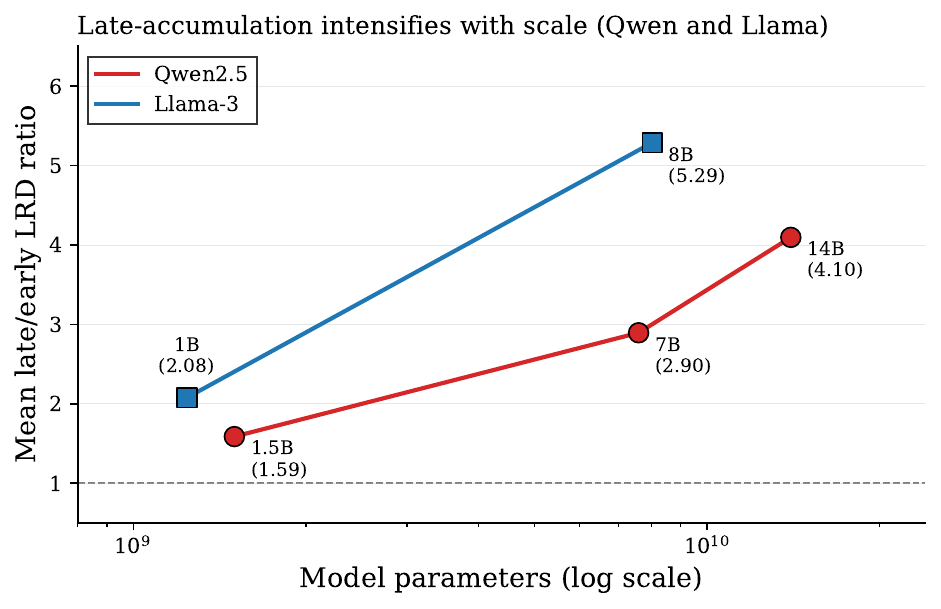}
\caption{Within-family scaling of late accumulation. Mean late/early
LRD ratio (deepest third over shallowest third, six-perturbation average)
vs.\ parameter count (log $x$). Qwen2.5: $1.59 \!\to\! 2.90 \!\to\!
4.10$; Llama: $2.08 \!\to\! 5.29$. Every per-perturbation cell is
monotonic within family (Table~\ref{tab:scaling}).}
\label{fig:qwen_scaling}
\end{figure*}

Qwen is treated as primary evidence and Llama as direction-only
corroboration; neither is a controlled multi-family sweep, and we
make no universal-scaling claim (Llama-3-8B uses $n{=}200$; its
late/early ratio is stable to ${\sim}0.2$ under bootstrap
resampling). A Mann--Whitney check on final-layer LRD (wrong vs.\
right) for Llama-3.2-1B-Instruct ($n{=}500$) matches the Llama-3-8B
per-perturbation pattern ($4$-of-$6$ significant).

\subsection{Per-Layer Cohen's $d$: Where the Failure Signal Emerges}
\label{sec:claim3}

If divergence causes failure, examples the perturbation breaks should
be separable from examples that survive---and the depth at which they
become separable marks where failures crystallize. We measure this
with per-layer Cohen's $d$: the standardized difference in
$\text{LRD}(L)$ between failing and succeeding examples
(\S\ref{sec:lora_method}), shown per model in
Appendix~\ref{sec:cohens_d_panels} (Figure~\ref{fig:supp_cohens_d}).
On Phi-3.5, $d$ is near-zero through layers 0--14 for most
perturbation types; succeed/fail separation develops at layers 20--27
and peaks at layers 28--32 ($d{\approx}0.4$--$0.6$; homophones
${\approx}0.9$), downstream of the activation-patching causal window. The same deep-layer concentration appears
on the four other models.
Two consequences follow: single-layer summaries are lossy
(final-layer LRD alone predicts failure weakly), and the early layers patching will
implicate (\S\ref{sec:claim4}) lie \emph{upstream} of where failures
crystallize.

\subsection{Activation Patching Locates the Causal Window}
\label{sec:claim4}

Single-layer clean patching (Appendix~\ref{sec:qwen14b_patching_extra},
Figure~\ref{fig:patching}) identifies the
causal window on the two confirmed-gate models: Phi-3.5
($n{=}50$~pairs, identity $100\%$) peaks at layers~1--8 and crosses
baseline at $L18$; Llama-3-8B ($n{=}100$, identity $90\%$) peaks at
$L1$ and stays above baseline through $L8$. Gemma-2-9B (identity
$60\%$), Qwen2.5-7B ($67\%$), and Mistral (identity recovery unstable
across patching configurations, at most $65\%$) all fall below the
$80\%$ gate and are exploratory.

\paragraph{Qwen at scale clears the patching gate.}
On Qwen2.5-14B (48-layer sweep, $84$ clean-correct pairs of $100$),
identity recovery rises to $85.0\%$, clearing the gate the $7$B
variant did not ($67\%$). The full curve
(Appendix~\ref{sec:qwen14b_patching_extra},
Figure~\ref{fig:supp_qwen14b_perlayer}) is consistent with
late-accumulation: early and mid thirds recover $73.1\%$ and $71.1\%$
of failed examples, while the late third falls to $42.3\%$. A
higher-than-expected random-patch control ($20.5\%$) weakens
effect-size attribution but not the identity-ceiling promotion; the
$14$B curve is a within-family consistency check, not a third entry
in the pre-registered two-model Spearman panel.

Note the tension already visible: patching implicates \emph{early}
layers, while the failure signal of \S\ref{sec:claim3} crystallizes
\emph{late}.

\subsection{Three-Map Dissociation}
\label{sec:threemap}

When the three per-layer signals---LRD, patching recovery, LoRA
effectiveness---are overlaid (Figure~\ref{fig:threemap_main}; all
five models in Appendix~\ref{sec:three_map_overlays},
Figure~\ref{fig:supp_three_map}), their peaks do not coincide, and
the relationships among them differ by regime.

\begin{figure*}[t]
\centering
\begin{subfigure}{0.46\textwidth}
  \centering
  \includegraphics[width=\linewidth]{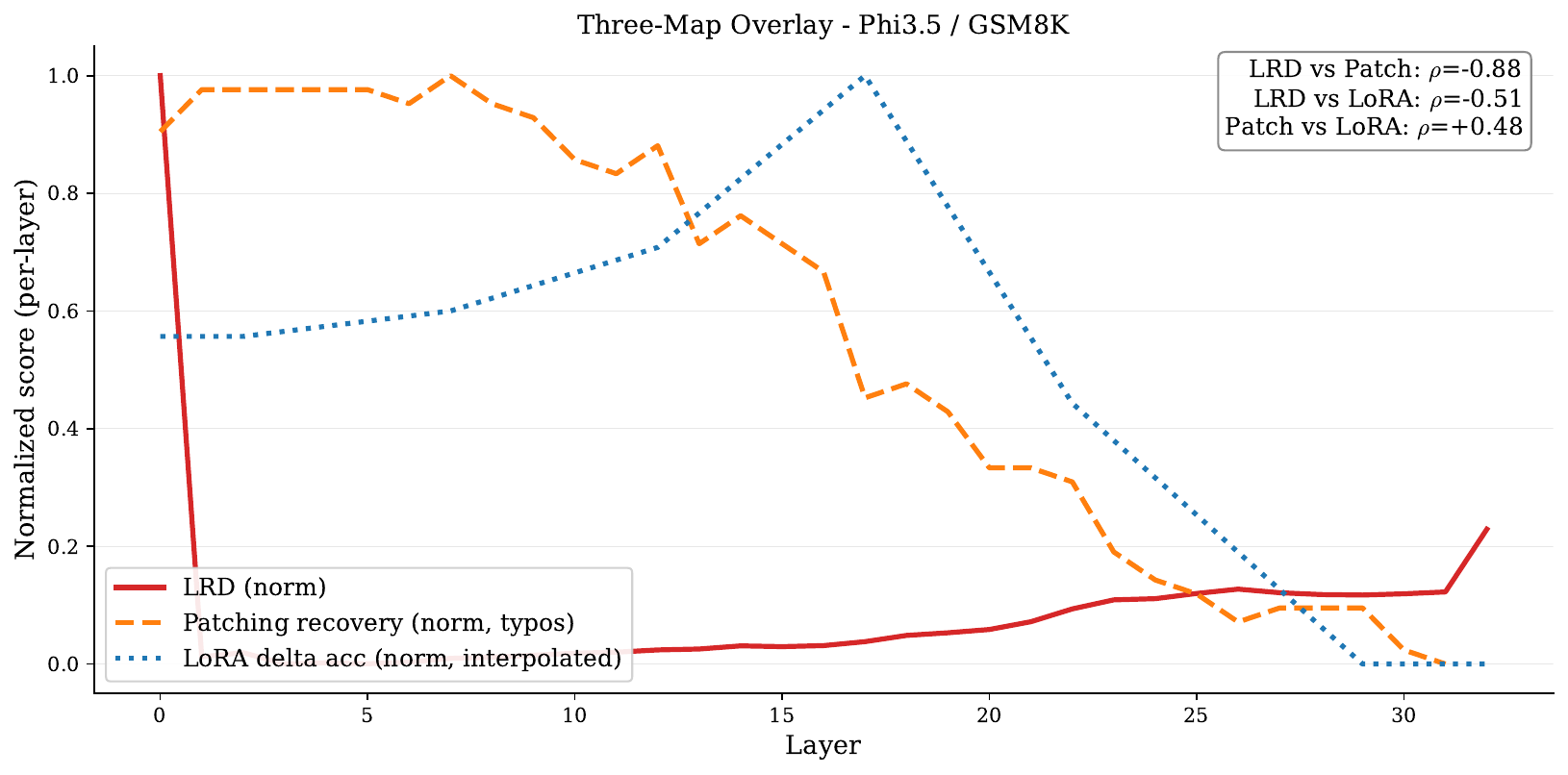}
  \subcaption{Phi-3.5 (spike-and-suppress): LoRA is most effective
  where LRD is suppressed.}
  \label{fig:threemap_main_phi}
\end{subfigure}\hfill
\begin{subfigure}{0.46\textwidth}
  \centering
  \includegraphics[width=\linewidth]{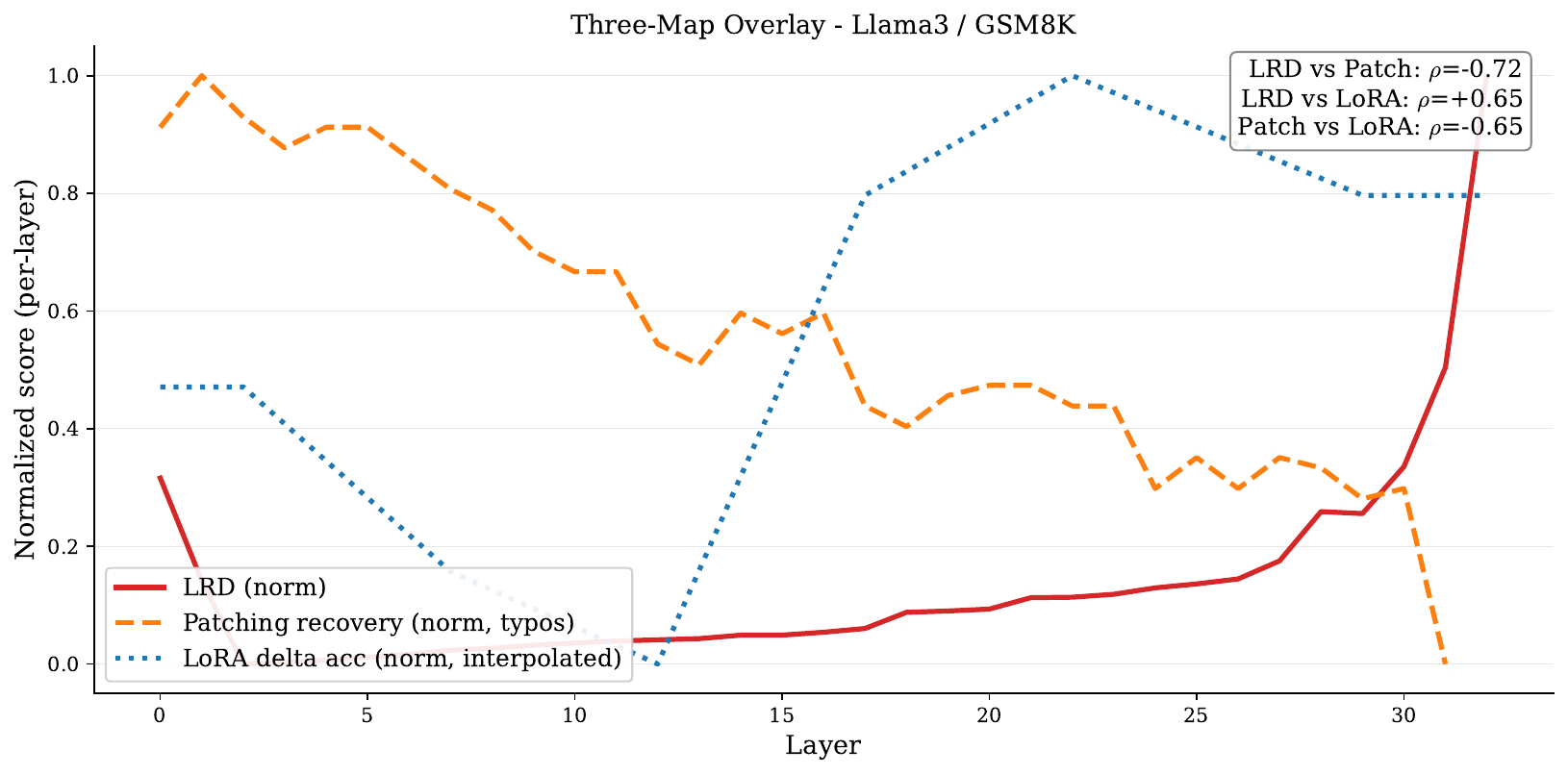}
  \subcaption{Llama-3 (late-accumulation): LoRA is most effective
  where LRD is still building.}
  \label{fig:threemap_main_llama}
\end{subfigure}
\caption{\textbf{The three maps dissociate.} Per-layer sensitivity
(LRD), causality (patching recovery), and compensatory capacity (LoRA
effectiveness), one model per propagation regime; the signals peak at
different layers, and sensitivity and causality are anti-correlated
on both models. Remaining models: Figure~\ref{fig:supp_three_map};
correlations: Table~\ref{tab:correlations}.}
\label{fig:threemap_main}
\end{figure*}

\paragraph{LRD and patching are anti-correlated.}
On the two confirmed-gate models (Phi-3.5, Llama-3),
$\rho = -0.72$ to $-0.88$, $p < 0.001$ (eff~$N \approx 3$--$5$). Mistral
matches the anti-correlation directionally but identity recovery falls
below the $80\%$ gate; we treat it as exploratory.
Gemma-2-9B is an exception ($\rho = +0.389$); we conjecture its
$60\%$ identity ceiling corrupts the patching signal, and because
Gemma fails the same gate, the exception sits outside the confirmed
panel rather than contradicting it.

\paragraph{LRD~$\times$~LoRA sign tracks regime.}
LoRA is most effective where LRD is suppressed (Phi-3.5, $\rho = -0.55$) or
still accumulating (Llama-3/Mistral, $\rho = +0.63$, $+0.69$). After
autocorrelation correction only Mistral reaches $p<0.05$ individually
($p=0.025$); we report the sign pattern across the panel rather than
per-model significance. Mistral's adapters here are from the same
shared-grid learning rate later diagnosed as lr-sensitive
(\S\ref{sec:tally}); we retain the correlation as directional evidence
only, consistent with Mistral's exclusion from formal sweep
adjudication.

\paragraph{C3/C4 intrinsic metrics are post-hoc only.}
The frozen-model intrinsic metrics---weight effective rank~(C3) and
gradient norm~(C4); \S\ref{sec:stats_method}---showed post-hoc
correlation with LoRA effectiveness on the original three models
($|\rho| = 0.54$--$0.75$, regime-matched signs) but failed prospective
validation: C3/C4 predicted L08--11 for Qwen and L24--29 for Gemma,
but the fixed-harness sweep (\S\ref{sec:sweep}) finds Qwen L08--11 to
be the \emph{worst} window ($-7.8$~pp) and Gemma flat at every
completed window. We treat C3/C4 as descriptive only
(full correlations and the prediction-vs-outcome summary:
Tables~\ref{tab:correlations} and~\ref{tab:supp_c3c4}).

\subsection{Cascade Disruption: Why Early Intervention Fails}
\label{sec:disruption}

If early-layer adapters fail because they break computation
\emph{downstream} of themselves, the damage should be directly
measurable: an adapted model should diverge from the frozen model on
\emph{clean} inputs, at layers after the adapter window. For each
trained LoRA window we compute the LRD between frozen and adapted
models on clean inputs at every downstream layer
(Appendix~\ref{sec:cascade_per_model}, Figure~\ref{fig:disruption}).

The measurements confirm the mechanism's signature: upstream of the
adapted window, hidden states are bit-identical to the frozen model
(divergence exactly zero); disruption onsets at the window, persists
through every downstream layer, and its total is largest for the
earliest windows, shrinking as the window moves deeper on all four
measured models (Phi-3.5, Qwen2.5-7B, Llama-3, Mistral;
Appendix~\ref{sec:cascade_per_model},
Figure~\ref{fig:supp_disruption}).
The rank order of total clean disruption across windows is inverse to
that of perturbed-accuracy gain (Appendix~\ref{sec:cascade_extra},
Figure~\ref{fig:supp_disruption_scatter}): placing LoRA at the layers
patching implicates breaks clean-input processing, so the
anti-correlation between the patching window and the best LoRA window
is mechanistic, not coincidental. (All profiles use the independently
retrained sweep adapters of Appendix~\ref{sec:cascade_extra}; both
quantities come from the same checkpoints.)

Cascade disruption makes a falsifiable behavioral prediction: the
windows the diagnostics flag should be the \emph{most} damaging
adapter sites, not the best. Before testing it
(\S\ref{sec:sweep}), we first check that the diagnostic maps
themselves are not artifacts of the task or the perturbation
distribution (\S\ref{sec:crosstask}).

\subsection{Cross-Task and Held-Out Generalization}
\label{sec:crosstask}

LRD profiles on Phi-3.5 and Mistral (the two models for which we
collected cross-task data) are near-identical across GSM8K, MMLU, and
BBH ($\rho > 0.95$ in both models): task modulates LRD magnitude but
not profile shape (Appendix~\ref{sec:crosstask_extra},
Figure~\ref{fig:supp_crosstask}). The regime signature is a property
of the model, not of the task on which it is measured.

\label{sec:heldout}
Nor is it a property of the perturbation distribution: we ran LRD on
three held-out perturbations (\texttt{char\_insert},
\texttt{char\_swap}, \texttt{qwerty}; absent from training and the
main analysis), one model per regime. Final-layer LRD on GSM8K:
Phi-3.5 (S\&S) $0.016$,
$0.023$, $0.020$; Llama-3-8B (Late) $0.066$, $0.105$, $0.082$---a
${\sim}5\times$ regime gap. Severity and error rate also dissociate:
Llama-3-8B has higher LRD yet loses fewer of its clean-correct
examples to the perturbation ($87$--$90\%$ survive, vs.\
$60$--$72\%$ for Phi-3.5; each scored on the model's own $n{=}200$
clean-correct subset), consistent with LRD measuring cascade severity
rather than error rate.

\subsection{How the Pieces Support the Dissociation Claim}
\label{sec:evidence}

The paper's central claim is that sensitivity, causality, and
compensatory capacity are dissociated per-layer properties. No single
experiment above is decisive; the claim rests on six convergent
strands, each ruling out a different deflationary explanation. (i)~\emph{LRD~$\times$~patching anti-correlation on both
confirmed-gate models} (\S\ref{sec:threemap}): if sensitivity and
causality were one underlying map read through noise---noisy
agreement---the correlation would be positive, not strongly negative.
This comparison alone kills that alternative outright.
(ii)~\emph{Regime-locked sign flip of LRD~$\times$~LoRA}
(\S\ref{sec:threemap}): the LoRA map's relationship to LRD flips sign
with propagation regime---the dissociation has regime structure.
(iii)~\emph{Cross-task profile stability} ($\rho > 0.95$ across
GSM8K/MMLU/BBH; \S\ref{sec:crosstask}) rules out a single-task
artifact. (iv)~\emph{Held-out perturbation types reproduce the regime
gap} (\S\ref{sec:heldout}), ruling out an artifact of the training
perturbation distribution. (v)~\emph{Layer-sweep asymmetry}
(\S\ref{sec:sweep}): diagnostic-flagged sites are the \emph{worst}
adapter windows---a behavioral confirmation that reuses none of the
correlational machinery. (vi)~\emph{Within-family scaling
monotonicity} (\S\ref{sec:scaling}) weakens the possibility that the
taxonomy merely relabels architectural family.

The weak links, stated plainly: the effective sample size behind each
correlation is only ${\approx}3$--$5$ after decorrelation; the
confirmed anti-correlation rests on the two gate-passing models (plus
Mistral directionally); and the LRD~$\times$~LoRA correlations mostly
miss per-model significance---we claim the sign pattern, not the
individual coefficients. The noisy-agreement alternative has to
survive all six strands simultaneously---strand~(i) alone already
rules it out, and each remaining strand closes off a different
deflationary explanation in turn.

\section{Layer Sweep: Diagnostic-Flagged Sites Resist Repair}
\label{sec:sweep}

The cascade-disruption result (\S\ref{sec:disruption}) predicts that
the patching-implicated early layers should be \emph{the worst}, not
the best, sites for perturbation-correction adapters. We test it by
sweeping adapter windows across the depth of Phi-3.5, Qwen2.5-7B,
Mistral, and Llama-3 ($3.8$--$8$B) plus the $9$B Gemma base, under
the fixed harness used throughout. Two of the five are not
adjudicable: Mistral (learning-rate artifact) and Gemma (adapter
insensitivity; \S\ref{sec:tally}).

\paragraph{Setup.} We train CE-only LoRA (configuration of
\S\ref{sec:lora_method}; $\lambda_{\text{stab}}{=}0$)
on perturbed GSM8K with clean supervision, sweeping non-overlapping
$5$-layer windows across the depth of each model with three seeds per
window. The perturbed
$\Delta$ is the mean change in perturbed
accuracy over the six perturbation types relative to the no-adapter
baseline, under the fixed harness (\S\ref{sec:negative}).

\subsection{Per-Model Layer Sweeps}
\label{sec:sweep_per_model}

\begin{table*}[t]
\centering
\small
\setlength{\tabcolsep}{3.5pt}
\begin{tabular}{lccccccc}
\toprule
Model (clean) & L00--04 & L05--09 & Early-mid & Mid & Mid-late & Late & All layers \\
\midrule
Phi-3.5 ($85.4\%$)         & $-1.4 \pm 0.7$  & $-0.9 \pm 0.1$  & $-5.3 \pm 1.1$  & $-3.5 \pm 0.8$  & $-0.1 \pm 0.2$  & $+0.2 \pm 0.3$ & $-8.6 \pm 0.7$ \\
Qwen2.5-7B ($89.0\%$)      & $-0.6 \pm 0.2$  & $-4.0 \pm 0.3$  & $-7.8 \pm 1.7$  & $-3.4 \pm 0.2$  & $-0.3 \pm 0.1$  & $+0.3 \pm 0.0$ & $-13.6 \pm 0.1$ \\
Llama-3-8B ($78.4\%$)      & $-2.5 \pm 0.2$  & $-6.4 \pm 0.5$  & ---             & $-2.7 \pm 0.8$  & $+0.4 \pm 0.2$  & $+1.1 \pm 0.1$ & $-9.9 \pm 1.0$ \\
Mistral-7B-v0.3 ($59.6\%$) & $-7.8 \pm 2.5$  & $-17.7 \pm 1.1$ & ---             & $-19.4 \pm 0.9$ & $-8.5 \pm 0.4$  & $-4.3 \pm 0.3$ & --- \\
Gemma-2-9B ($64.8\%$)$^{s}$& $+0.2$          & $-0.2$          & ---             & $+0.5$          & $+0.2$          & $0.0$          & --- \\
\bottomrule
\end{tabular}
\caption{Mean perturbed $\Delta$ (with-adapter minus no-adapter, six
perturbations on $500$ GSM8K items); mean~$\pm$~sample std over three
seeds ($^{s}$: Gemma single-seed). Headers are nominal; exact windows,
the Gemma protocol, and the pre-registered Qwen L08--11 cell (C3/C4's
pick---Qwen's worst window) are in
Appendix~\ref{sec:layer_sweep_per_model}. Dashes: not trained.
\emph{Mistral excluded from adjudication (lr artifact;
\S\ref{sec:tally}).} ``All layers'' adapts every layer; it regresses
more than any single window on all three adjudicable models.}
\label{tab:layer_sweep_panel}
\end{table*}

Table~\ref{tab:layer_sweep_panel} gives the full grid;
Figure~\ref{fig:layer_sweep} plots the sweep, and per-model panels
with seed dispersion are in Figure~\ref{fig:supp_layer_sweep_per_model}
(both in Appendix~\ref{sec:layer_sweep_per_model}).

\paragraph{Per-model results.}
On Phi-3.5 (S\&S; $32$ layers) every mid-layer window is strongly
negative; the only non-negative window is L27--31 ($+0.2$~pp,
effectively null). Seed std (sample, $n{-}1$) ranges $0.1$--$1.1$~pp,
no window crosses zero under seed variation, and the earlier-harness
L15--19 ``optimum'' falls to $-3.5$ under the fixed harness.
Qwen2.5-7B's
worst window is the C3/C4-predicted L08--11 ($-7.8$~pp); its
least-regressive is the deepest, L24--27 ($+0.3$). Llama-3-8B has
L20--24 and L27--31 non-negative ($+0.4$, $+1.1$). Mistral-7B-v0.3 is
uniformly negative at the shared lr; a single-cell audit at
lr$=1\mathrm{e}{-}5$ recovers to $+0.13$~pp, consistent with an lr
artifact (\S\ref{sec:tally}). Gemma-2-9B is adapter-insensitive
(every window within $\pm 0.7$~pp of zero) and we treat it as ``no
measurable signal.''

\subsection{The Mid-Layer Wall and the Pre-Registered Placement Tally}
\label{sec:tally}

Two structural observations follow from
Table~\ref{tab:layer_sweep_panel}. First, \textbf{mid-layer windows
uniformly backfire on chain-of-thought GSM8K} (sign-consistent but
strongly attenuated on short-form MMLU; see below): on all three
adjudicable models, every window in the L05--L20 range is strongly
negative, and the patching-implicated early windows (L00--04) are
also negative, though less catastrophically. Second,
\textbf{least-regressive windows are the deepest ones}: the only
non-negative windows on the three models are the deepest available,
confirming the cascade-disruption prediction that placement should be
\emph{away from} where divergence and patching causality concentrate.

\paragraph{Adjudicated tally.}
Excluding Mistral (lr artifact) and Gemma (flat at every window), the
pre-registered prediction was that windows away from the
patching-implicated early layers would be less damaging than the
flagged layers. In addition, C3/C4---fit post hoc on the original
three models---was registered prospectively for Qwen-7B before its
sweep ran, marking L08--11 as its best window.
The tally: Qwen \checkmark{} (L24--27 beats L08--11 by $8.1$~pp, and
the C3/C4 pick is Qwen's \emph{worst} window); Llama-3 \checkmark{}
(L27--31 beats L05--09 by $7.5$~pp); Phi-3.5 $\times$ (scored as a
fail: no window rises meaningfully above zero, though the relative
ordering---mid worst, deepest least regressive---matches). Two holds
and one fail. What the sweep delivers is a \emph{validated placement
anti-pattern with a mechanism}: do not place perturbation-repair
adapters where LRD or patching point. It is a relative-ranking rule,
not a positive recipe---absolute gains over the no-adapter baseline
are small (Llama-3 $+1.1$~pp) or effectively null (Phi-3.5, Qwen),
and Phi-3.5's flatness is why its tally entry is a fail.

\paragraph{Width robustness and dose response.}
The wall is not an artifact of the fixed $5$-layer window: a
window-width ablation on Phi-3.5 (widths $3$/$5$/$7$;
Appendix~\ref{sec:width_mmlu}, Table~\ref{tab:width_ablation})
preserves the sign structure at every width, and mid-window damage
grows monotonically with width ($-2.8 \to -5.3 \to -7.8$~pp). The
all-layer baseline (Table~\ref{tab:layer_sweep_panel}, last column)
extends this dose response to its endpoint: adapting every layer
regresses perturbed accuracy more than any single window on all three
adjudicable models, and regresses \emph{clean}
accuracy as well ($-8.1$ to $-14.1$~pp; Appendix~\ref{sec:width_mmlu}),
as expected if the early-layer share of an all-layer adapter incurs
cascade damage. Relative to all-layer adaptation, the deepest window
is $9$--$14$~pp less regressive while training a fraction of the
parameters.

\paragraph{Beyond chain-of-thought: an MMLU control.}
A fixed-harness sweep on MMLU (multiple-choice; one model per regime;
Appendix~\ref{sec:width_mmlu}, Table~\ref{tab:mmlu_sweep}) is
sign-consistent with the GSM8K wall but strongly attenuated---we do
not claim the wall replicates there. The only reliably nonzero cell is
Phi-3.5's diagnostic-flagged mid window L10--14 ($-1.8 \pm 0.4$~pp);
no window on either model produces positive transfer, so the absence
of a positive placement recipe holds on both tasks. The attenuation is
consistent with cascade disruption, which predicts damage compounds
over autoregressive generation steps---long chain-of-thought
generations should suffer far more than single-token multiple
choice---an exploratory interpretation.
MMLU's short-answer format also makes generation-length harness
artifacts (\S\ref{sec:negative}) structurally impossible, giving the
sweep an artifact-immune task.

\subsection{An Attempted Intervention and a Harness Artifact}
\label{sec:negative}

We explored the representation-stability loss $L_{\text{stab}}$ of
\S\ref{sec:lora_method} added to CE LoRA. An earlier (pre-fix)
harness with \texttt{max\_new\_tokens}${=}100$ reported $+7.3$~pp on
Phi-3.5 at L15--19 and a regime-dependent split. Under the fixed harness those
gains reverse (Appendix~\ref{sec:earlier_investigations},
Figure~\ref{fig:harness_artifact};
\texttt{max\_new\_tokens}${=}512$ for this checkpoint comparison and
the layer sweep of \S\ref{sec:sweep}; settings in
Appendix~\ref{sec:repro}): Phi-3.5 L15--19 records $-3.5$~pp,
and the Qwen2.5-7B L24--27 vanilla CE baseline reads $-7.5$~pp where
the same checkpoint had read $+11.6$~pp. (The $-7.5$ re-scores the
\emph{earlier pipeline's} checkpoint; Table~\ref{tab:layer_sweep_panel}'s
$+0.3$ for the same window comes from adapters trained afresh.) The
cause is mundane: truncated chain-of-thought was scored as empty,
crediting conditions that produce shorter (often less accurate)
output.

The reversal is not merely bookkeeping. Holding the fixed harness
constant and toggling only $\lambda_{\text{stab}}$, the stability
term ($\lambda_{\text{stab}}=1.0$) \emph{hurts} perturbed accuracy by
$1.2$--$3.9$~pp across seeds on Phi-3.5 L15--19---the window where
the earlier harness reported $+7.3$~pp. The diagnostic, mechanism, and scaling results in
\S\ref{sec:diagnostics} are independent of this claim; the
reversal rules out a regime-dependent stabilizer benefit, leaving
cascade disruption---which predicts mid-layer harm---as the surviving
explanation.

\section{Discussion}
\label{sec:discussion}

\paragraph{Restoration vs.\ adaptation.}
Why should the causal map mislead? Patching substitutes the
\emph{exact} clean representation at inference time---an oracle
intervention---so it identifies where \emph{restoration} helps. An
adapter must instead \emph{learn} a correction applied to every
input, including clean ones, with low-rank weights. Early layers are
where restoration is
cheapest but also where a learned
modification propagates through the most downstream computation; the
layer sweep shows that cost dominates, damage scaling monotonically
with adapted-window width (\S\ref{sec:tally}).

\paragraph{Implications for practitioners.}
First, LRD is a cheap pre-screen: forward passes on clean/perturbed
pairs identify a model's propagation regime---and rule out early- and
mid-layer adapter placement---before any training compute is spent.
Second, diagnostic-guided placement is falsified in this setting:
neither activation patching nor intrinsic weight metrics (C3/C4)
select good adapter sites, and both actively select bad ones; absent
task-specific evidence, default to the deepest window (a relative
ranking; absolute gains remain small or null). Third, any
intervention evaluated on
chain-of-thought tasks should ship with a generation-length-matched
null; re-scoring checkpoints under an adequate budget is nearly free.

\paragraph{Open questions.}
Regime initially tracked family within our panel; the within-family
scans weaken that confound but do not eliminate it, and we make no
universal-scaling claim. Two follow-ups: a controlled multi-family
scan varying architecture and scale independently, and a sweep across
tasks of intermediate generation length to test the
compounding-damage account directly.

\section{Conclusion}
\label{sec:conclusion}

Which layer is ``responsible'' for a perturbation failure depends on
what responsible means: sensitivity, causality, and compensatory
capacity give three different per-layer answers, and the disagreement
is systematic. The regimes are stable across tasks and perturbation
types and strengthen with scale within two families. Cascade
disruption explains why causally implicated layers
resist repair; a fixed-harness sweep confirms its central
prediction---diagnostic-flagged sites are the most damaging adapter
windows---on every adjudicable model---and an intervention we once
believed in did not survive re-evaluation under an adequate
generation budget.

\section*{Limitations}

\paragraph{Scope of intervention claims.}
We provide no positive intervention recipe. All placement-related
claims are relative-ranking statements under a fixed harness (``window
$A$ is less regressive than window $B$''); absolute gains over
no-adapter baselines are not claimed. The earlier-pipeline
representation-stability gains were a harness artifact
(\S\ref{sec:negative}); the underlying cascade-disruption mechanism is
independently supported by the clean-input disruption measurements
(\S\ref{sec:disruption}).

\paragraph{Panel size.}
Our regime taxonomy is observed on a five-model panel. The primary
de-confounding evidence is the three-point Qwen2.5 scan
(\S\ref{sec:scaling}); the two-point Llama scan corroborates the
direction on a second family but has a narrow-bin caveat at the $1$B
end ($17$ hidden layers). Two families is not a controlled
multi-family sweep, and confirming regime as a general transformer
property still requires within-family scans on additional families
and, ideally, controlled pretraining recipes.

\paragraph{Task and generation-length scope.}
The layer-sweep adjudication uses GSM8K, with MMLU as a short-form
control on one model per regime (\S\ref{sec:tally}). Mid-layer damage
magnitude appears generation-length dependent: on MMLU the only
reliably nonzero window is Phi-3.5's mid window ($-1.8$~pp) and all
other windows on both models are within noise, so effects on
short-form and multiple-choice tasks are strongly attenuated---tested
on MMLU only, with no sweep at intermediate generation lengths.
Cross-task LRD stability is measured on two models (Phi-3.5 and
Mistral; \S\ref{sec:crosstask}) across GSM8K/MMLU/BBH. All
experiments are English-only.

\paragraph{Patching ceiling effects.}
Gemma-2-9B and Qwen2.5-7B identity ceilings ($60\%$, $67\%$) limit
causal interpretation when perturbation effects distribute across
layers. Mistral's identity recovery varies across patching
configurations (at most $65\%$, with large run-to-run variance), so
we report no single ceiling for it.
Qwen-14B clears the $80\%$ gate ($85.0\%$) but with a $20.5\%$
random-patch warning that weakens effect-size attribution on the $14$B
curve.

\paragraph{Statistical power.}
Three seeds per layer-sweep window characterizes each cell's direction
and magnitude but limits the precision of fine-grained between-window
comparisons. Activation-patching estimates likewise rest on small pair
counts (Phi-3.5 $n{=}50$, Llama-3 $n{=}100$, Qwen-14B $84$ pairs). The main sweep uses a fixed $5$-layer window; the
width ablation (\S\ref{sec:tally}, Table~\ref{tab:width_ablation})
shows the sign structure is width-robust at $3$/$5$/$7$, but covers
one model (Phi-3.5).
The block-bootstrap correlation analyses use block size
$5$ from the partial-autocorrelation decay; the post-correction effective
$N{\approx}3$--$5$ is already small. A block-size sensitivity sweep
($b{=}2$--$8$; Appendix~\ref{sec:block_sensitivity}) shows the primary
LRD-vs-patching anti-correlation stays significant ($p<0.05$) at every
$b$ for all three models, while the LRD-vs-LoRA and patching-vs-LoRA
pairs are not uniformly significant across $b$---consistent with the
sign-pattern-not-coefficients framing above.

\paragraph{Mistral learning-rate sensitivity.}
At the shared-grid lr ($5\mathrm{e}{-}5$) every Mistral window is
strongly negative; a single-cell re-run at $1\mathrm{e}{-}5$ recovers
to $+0.13$~pp, consistent with an lr artifact rather than a capacity
issue, though we did not re-run every window. Mistral is excluded
from the placement-prediction adjudication.

\paragraph{Code and data.}
Code, configurations, and per-model evaluation logs are available at
\url{https://github.com/NathanLabiosa/Layer-Wise-Representation-Divergence}.

\paragraph{Use of AI assistants.}
AI coding assistants were used during this project for code
scaffolding, experiment orchestration (job-launch scripts, log
parsing, plot generation), and writing assistance (copy-editing and reorganization of paper text). All scientific
claims, experimental designs, statistical analyses, and result
interpretations are the authors' own; AI-generated text and code
were reviewed and verified by the authors before inclusion.

\paragraph{Artifact use and licensing.}
All models and benchmarks used here are publicly released research
artifacts, cited at first use in \S\ref{sec:diagnostics} and listed in
references. Licenses, to the best of our knowledge at the time of
writing: Phi-3.5-mini-instruct (MIT), Mistral-7B-Instruct-v0.3
(Apache~2.0), Llama-3-8B-Instruct and Llama-3.2-1B-Instruct (Llama-3
Community Licenses), Gemma-2-9B (Gemma Terms of Use),
Qwen2.5-\{$1.5$B, $7$B, $14$B\}-Instruct (Apache~2.0), GSM8K (MIT),
MMLU (MIT), BBH (Apache~2.0). Our usage---measuring representational
properties and training diagnostic LoRA adapters for academic
research---is consistent with the intended use stated by each
artifact's release. The benchmarks (GSM8K, MMLU, BBH) consist of
grade-school math word problems, multiple-choice exam questions, and
reasoning challenges respectively, and to our knowledge contain no
personally identifying information or offensive content; each is
documented in its release paper (cited).

\paragraph{Potential risks.}
This work is diagnostic and does not release new models, training
data, or adversarial tools. The perturbation types we study (typos,
OCR noise, whitespace, case, speech-like substitutions, homophones)
are commonplace and not novel attack vectors. The diagnostic itself
(LRD, patching recovery, layer sweeps) could in principle inform
adversarial perturbation design or targeted intervention against
deployed models; we judge this risk low because the techniques and
perturbations are already public. The more practical risk is
\emph{misuse of the regime taxonomy}: extrapolating the
spike-and-suppress / late-accumulation labels to models outside our
five-model panel without re-running the diagnostic, or treating the
post-hoc C3/C4 metrics as a placement rule despite the falsification
in \S\ref{sec:threemap}. The Limitations above are written to make
both kinds of over-reach explicit. Finally, the layer-sweep and
diagnostic experiments are GPU-intensive (multi-model LoRA sweeps
with three seeds per window, plus activation patching at every
layer); we report exact configurations in Appendix~\ref{sec:repro} so
practitioners can scope re-runs to the cells they need.

\bibliography{references}

\clearpage
\appendix
This appendix provides extended results referenced in the main paper.
\S\ref{sec:lrd_extra} contains LRD heatmaps for all five panel models,
and \S\ref{sec:scaling_extra} adds Qwen-1.5B and Qwen-14B heatmaps
that complement the within-family scaling analysis
(Figure~\ref{fig:qwen_scaling}, \S\ref{sec:scaling}).
\S\ref{sec:qwen14b_patching_extra} gives the full $48$-layer
Qwen2.5-14B patching curve summarized in \S\ref{sec:claim4}.
\S\ref{sec:per_pert}--\S\ref{sec:intrinsic} give per-perturbation
recovery rates, the full panel of pairwise Spearman correlations across
all five models, and the intrinsic-predictor (C1--C4) definitions and
C3/C4 falsification table.
\S\ref{sec:three_map_overlays}--\S\ref{sec:cascade_per_model} contain
per-model figure panels for the three-map overlay (all five models),
Cohen's $d$ (all five models), intrinsic capacity metrics
(Llama-3/Mistral), and cascade disruption (Llama-3/Mistral).
\S\ref{sec:cascade_extra}--\S\ref{sec:crosstask_extra} provide the
disruption-vs-gain scatter (Phi-3.5, Llama-3, Mistral) and the
cross-task LRD profiles (Phi-3.5, Mistral).
\S\ref{sec:layer_sweep_per_model} contains the per-model layer-sweep
panels underlying Table~\ref{tab:layer_sweep_panel}.
\S\ref{sec:earlier_investigations} briefly catalogs investigations
performed before the evaluation harness was fixed---the cosine vs.\ MSE
stability-loss matched-rate controls, the $\lambda_{\text{stab}}$
scaling sanity-check, the LoRA capacity experiments, and a
residual-adapter stabilizer tested on Mistral and Qwen2.5-7B---and
explains why those numbers are excluded from this revision.
\S\ref{sec:repro} lists reproducibility settings for the fixed-harness
runs.

\section{LRD Profiles: All Five Panel Models}
\label{supp:lrd_profiles}
\label{sec:lrd_extra}

\begin{figure*}[!htbp]
\centering
\includegraphics[width=.80\textwidth]{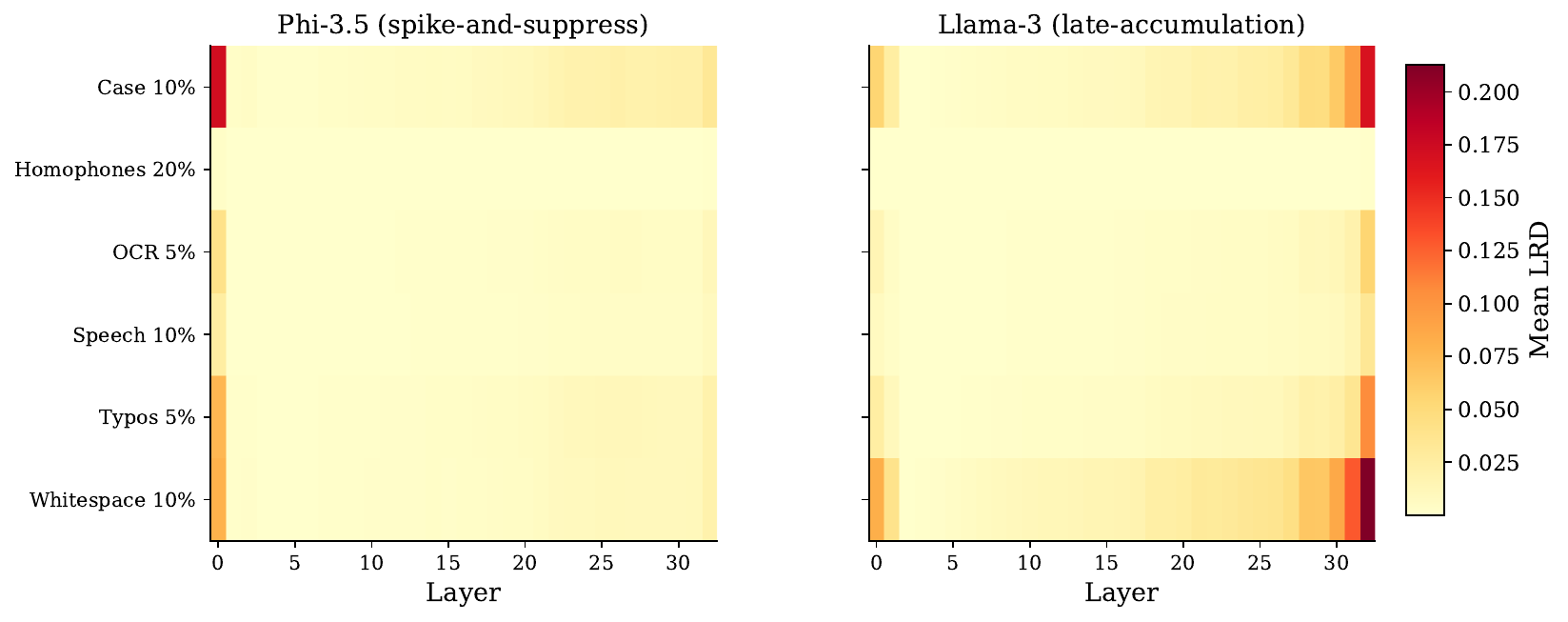}
\caption{LRD heatmaps (GSM8K, typos 5\%). Phi-3.5 shows
spike-and-suppress; Llama-3 shows late-accumulation. Color encodes LRD
magnitude on a shared scale; rows are layers, columns are examples.}
\label{fig:heatmap}
\end{figure*}

\begin{figure*}[!htbp]
\centering
\begin{subfigure}{0.62\textwidth}
  \centering
  \includegraphics[width=\linewidth]{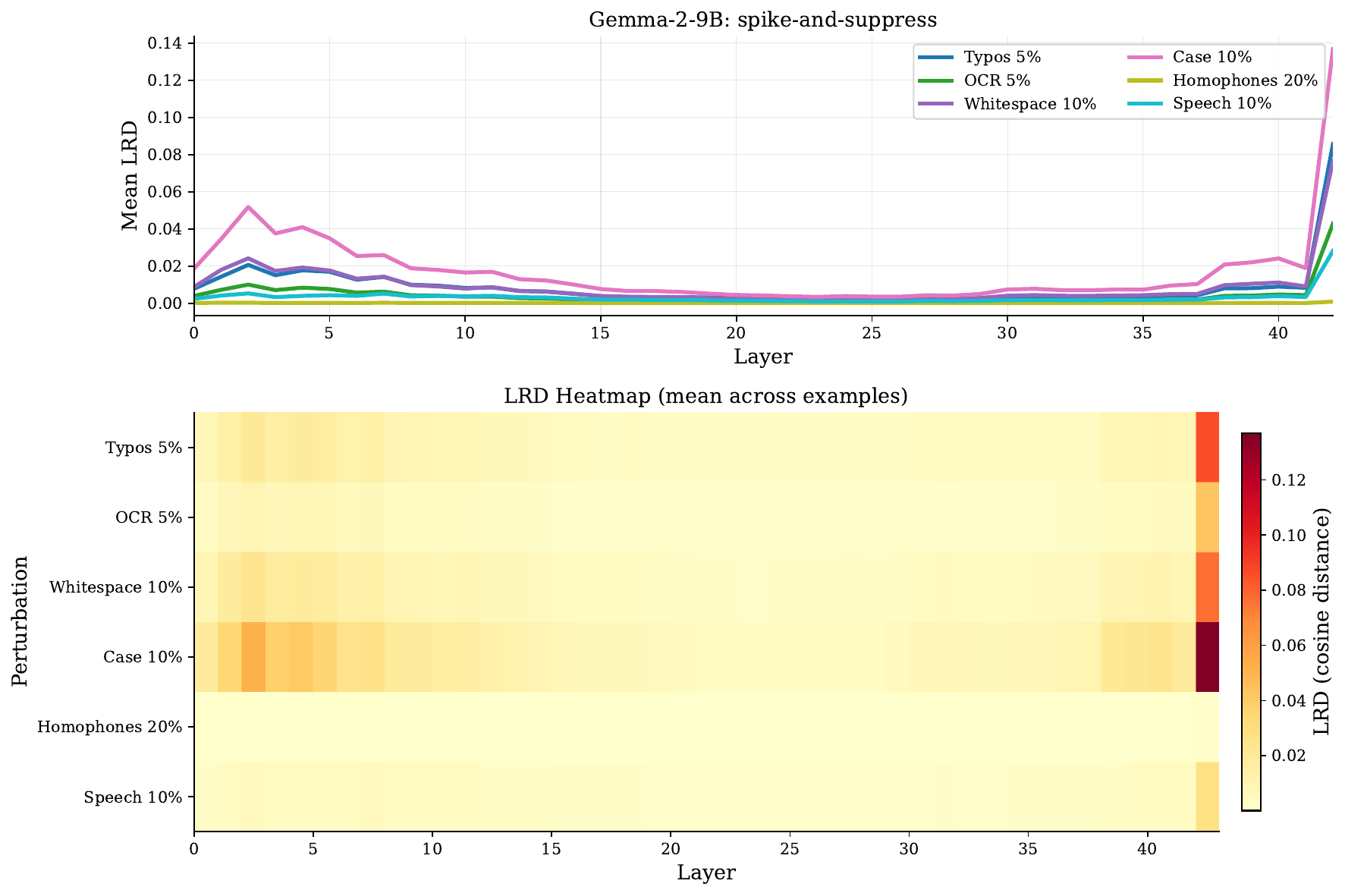}
  \subcaption{Gemma-2-9B. Spike-and-suppress profile with a mild secondary
  hump at layers~30--41 and a sharp jump at the final layer.}
  \label{fig:supp_lrd_gemma}
\end{subfigure}
\\[0.5em]
\begin{subfigure}{0.62\textwidth}
  \centering
  \includegraphics[width=\linewidth]{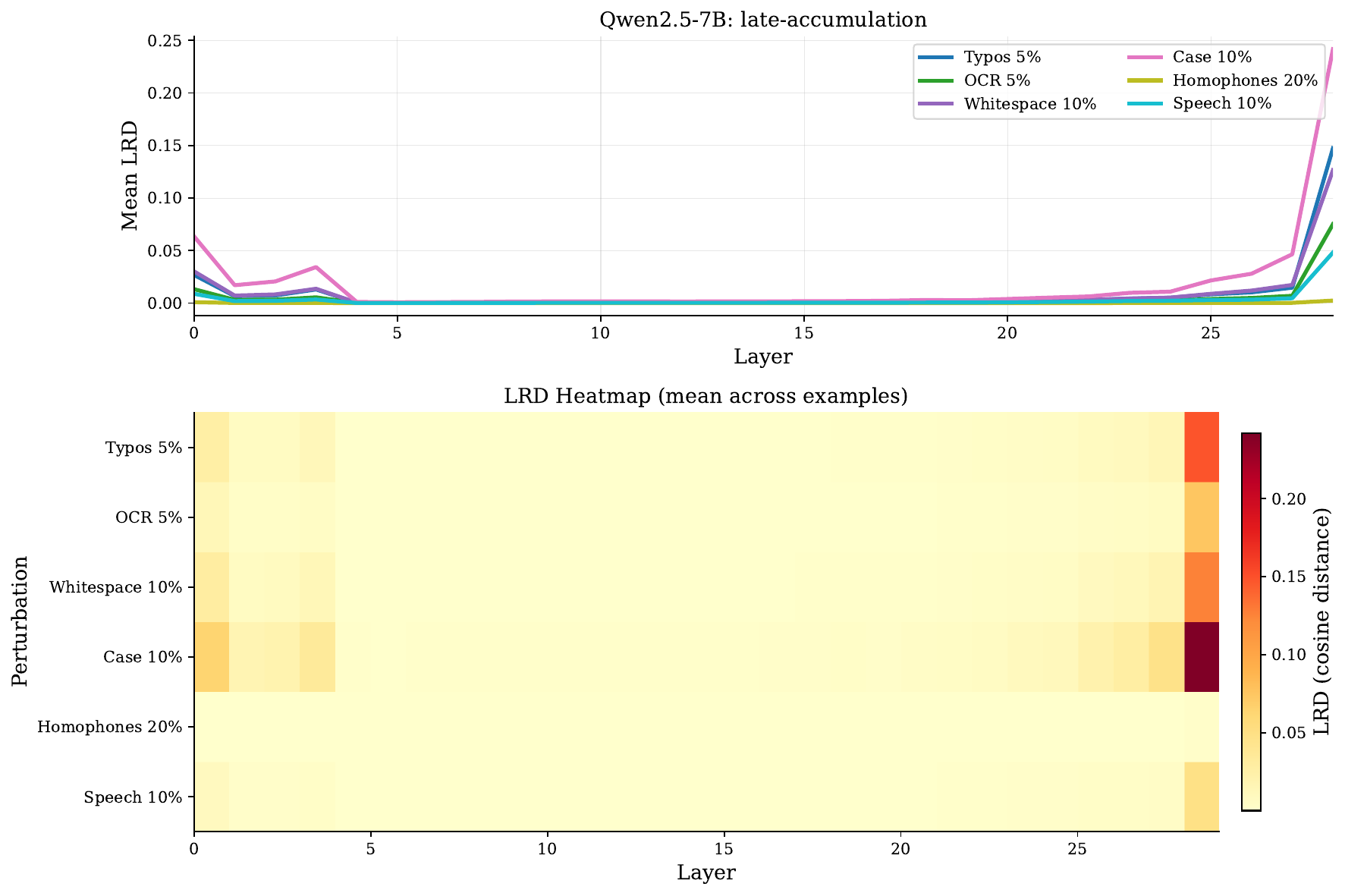}
  \subcaption{Qwen2.5-7B. Late-accumulation profile with an early transient
  spike at layers~0--3 and dominant late buildup through layers~20--28.}
  \label{fig:supp_lrd_qwen}
\end{subfigure}
\\[0.5em]
\begin{subfigure}{0.62\textwidth}
  \centering
  \includegraphics[width=\linewidth]{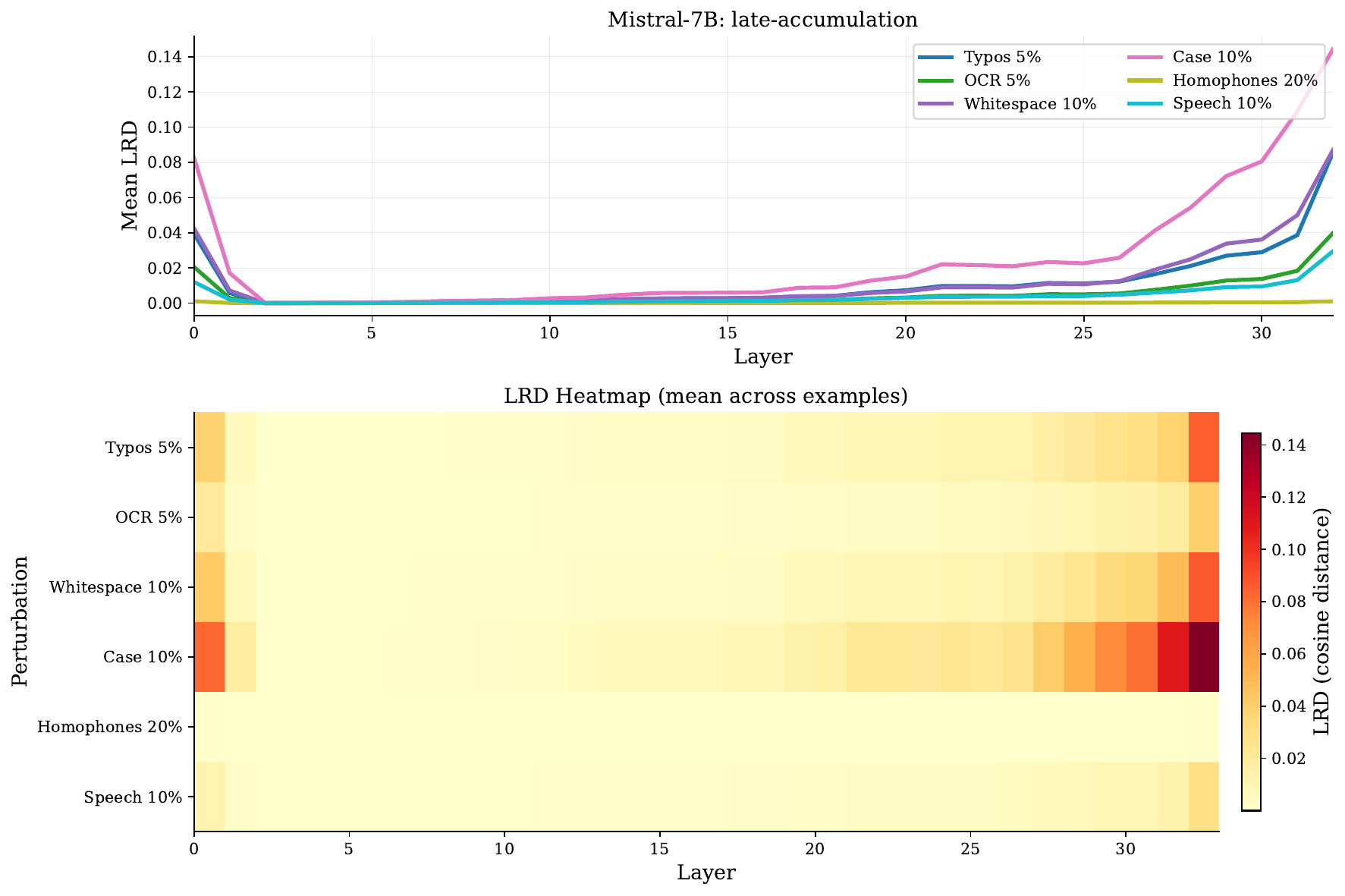}
  \subcaption{Mistral. Late-accumulation profile; gradual buildup through depth.}
  \label{fig:supp_lrd_mistral}
\end{subfigure}
\caption{LRD heatmaps (GSM8K, typos 5\%) for the three additional models.
Compare Phi-3.5 and Llama-3 in Figure~\ref{fig:heatmap}.}
\label{fig:supp_lrd}
\end{figure*}

Figure~\ref{fig:heatmap} shows LRD heatmaps for the two models that
anchor the regime taxonomy (\S\ref{sec:claim1}): Phi-3.5
(spike-and-suppress) and Llama-3 (late-accumulation).
Figure~\ref{fig:supp_lrd} shows the remaining three models. Gemma-2-9B
exhibits the spike-and-suppress profile (large early divergence,
suppressed through the middle layers, with a mild secondary hump near
the final layers). Qwen2.5-7B exhibits the late-accumulation profile,
with an early transient spike at layers~0--3 and dominant late buildup
through layers~20--28. Mistral shows a gradual late-accumulation
buildup through depth.

\section{Within-Family Scaling: Qwen-1.5B and Qwen-14B Heatmaps}
\label{sec:scaling_extra}

\begin{figure*}[!htbp]
\centering
\begin{subfigure}[t]{0.5\textwidth}
  \centering
  \includegraphics[width=\linewidth]{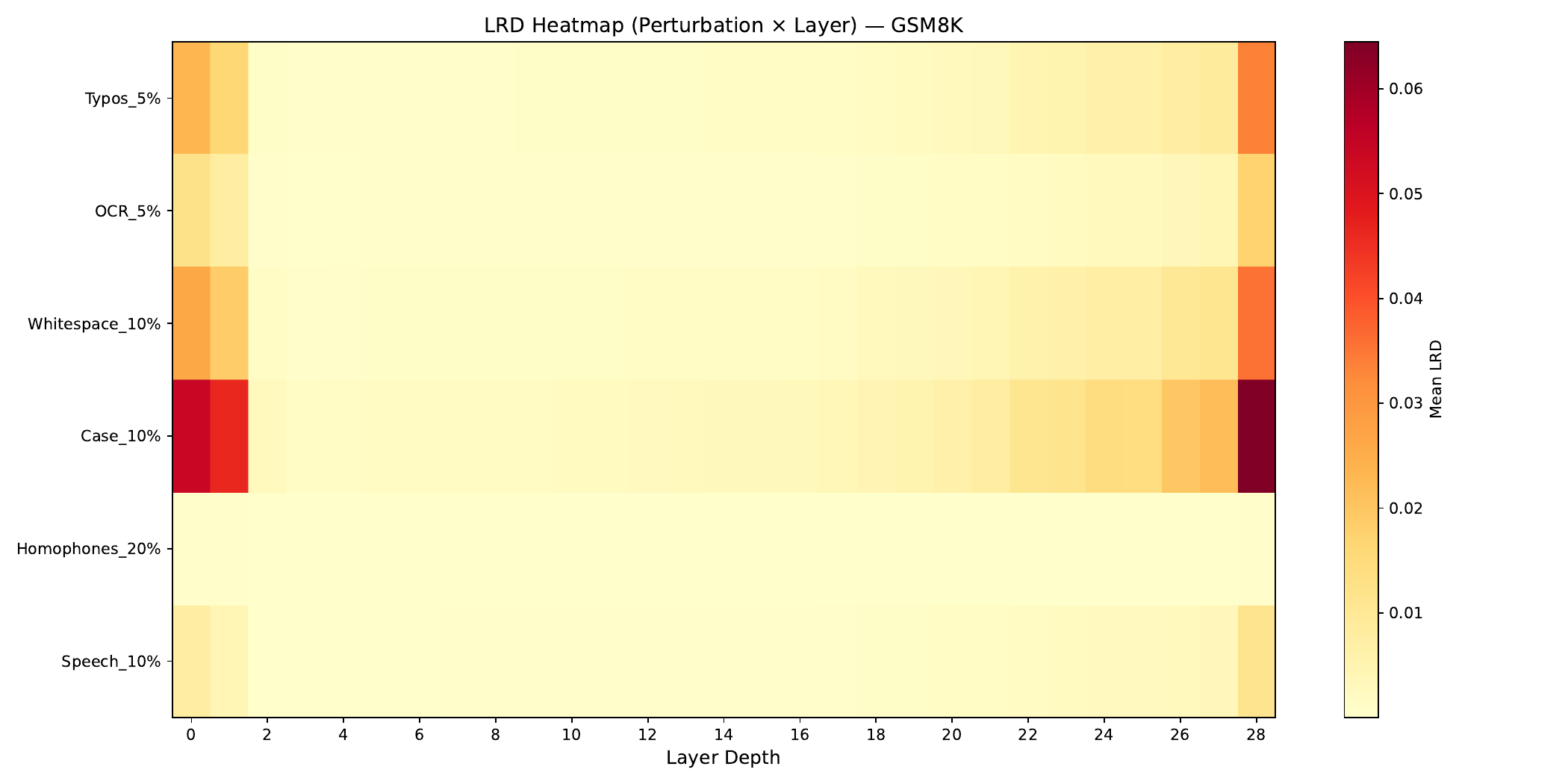}
  \caption{Qwen2.5-1.5B (late/early ratio $1.59$). Late-accumulation
  profile is present but mild; depth-driven divergence is visible after
  roughly the midpoint of layers.}
  \label{fig:supp_qwen_1p5b_heatmap}
\end{subfigure}
\\[0.8em]
\begin{subfigure}[t]{0.72\textwidth}
  \centering
  \includegraphics[width=\linewidth]{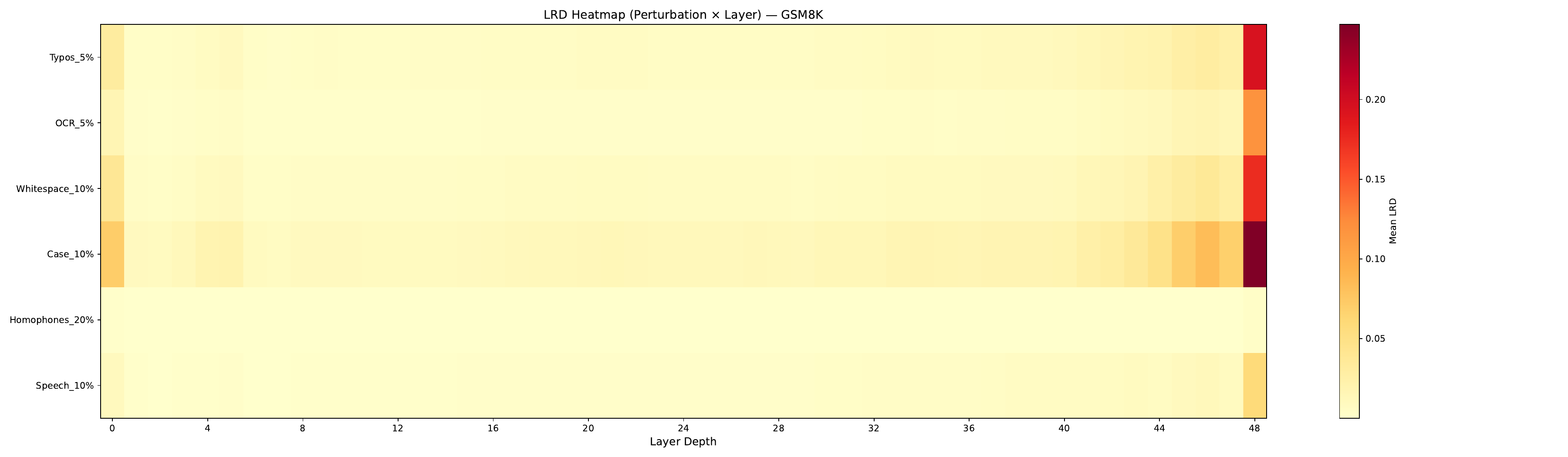}
  \caption{Qwen2.5-14B (late/early ratio $4.10$). Late-accumulation
  profile is sharper: deep-layer divergence dominates earlier and on
  every perturbation type.}
  \label{fig:supp_qwen_14b_heatmap}
\end{subfigure}
\caption{LRD heatmaps (GSM8K, typos~5\%) for Qwen2.5-1.5B and
Qwen2.5-14B. Compare with Qwen2.5-7B in
Figure~\ref{fig:supp_lrd_qwen}. The late-accumulation regime persists at
all three scales and intensifies monotonically with size.}
\label{fig:supp_scaling_heatmaps}
\end{figure*}

Figure~\ref{fig:supp_scaling_heatmaps} shows LRD heatmaps for Qwen-1.5B
and Qwen-14B; the Qwen-7B heatmap is in
Figure~\ref{fig:supp_lrd_qwen}. Together with the within-family scaling
figure in the main paper (Figure~\ref{fig:qwen_scaling}), these visualize the same
late-accumulation regime at three Qwen scales: an early transient spike
near layers~0--3 that fades, followed by a depth-dominant build-up
whose late/early ratio intensifies monotonically with scale
($1.59 \to 2.90 \to 4.10$). Table~\ref{tab:scaling} gives the
per-perturbation grid behind the main-paper scaling figure
(Figure~\ref{fig:qwen_scaling}, \S\ref{sec:scaling}).

\begin{table*}[!htbp]
\centering
\small
\begin{tabular}{llcccccc|c}
\toprule
Family & Model & Typos & OCR & Whitespace & Case & Homophones & Speech & Mean \\
\midrule
Qwen2.5 & 1.5B  & $1.65$ & $1.53$ & $1.68$ & $1.42$ & $1.34$ & $1.92$ & $\mathbf{1.59}$ \\
        & 7B    & $3.19$ & $3.40$ & $2.66$ & $2.40$ & $2.24$ & $3.49$ & $\mathbf{2.90}$ \\
        & 14B   & $4.20$ & $4.88$ & $3.55$ & $3.17$ & $3.84$ & $4.93$ & $\mathbf{4.10}$ \\
\midrule
Llama   & 3.2-1B & $2.44$ & $2.02$ & $2.24$ & $2.51$ & $1.28$ & $1.98$ & $\mathbf{2.08}$ \\
        & 3-8B$^{\S}$ & $5.90$ & $5.86$ & $4.50$ & $5.24$ & $3.25$ & $6.99$ & $\mathbf{5.29}$ \\
\bottomrule
\end{tabular}
\caption{Late/early LRD ratio per perturbation for the two scaling
scans; Mean averages the six entries. Every per-perturbation cell is
monotonic within family. GSM8K, $n{=}500$,
\texttt{max\_new\_tokens}${=}768$; $\S$~Llama-3-8B used $n{=}200$.}
\label{tab:scaling}
\end{table*}

\section{Activation-Patching Recovery Curves}
\label{sec:qwen14b_patching_extra}

\begin{figure*}[!htbp]
\centering
\includegraphics[width=0.75\textwidth]{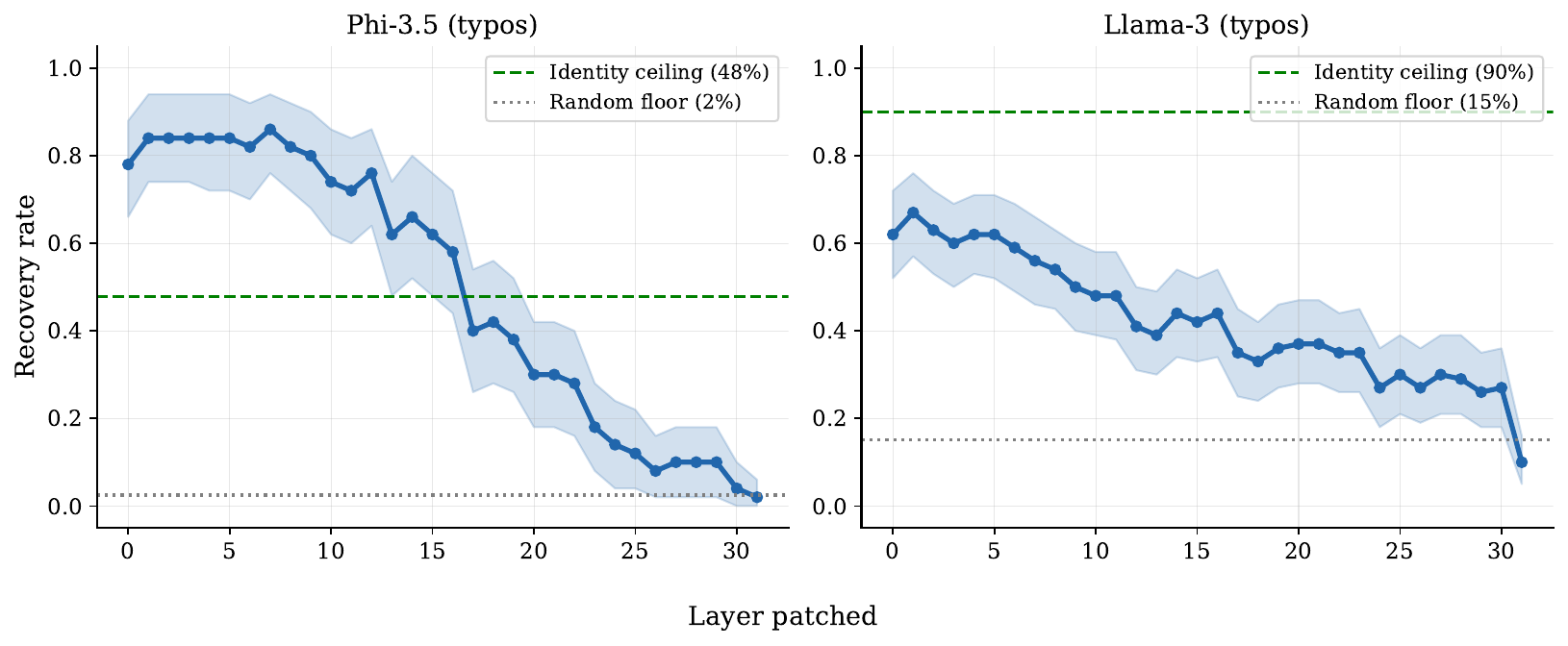}
\caption{Activation patching recovery on the two confirmed-gate
models (Phi-3.5, Llama-3; $3.8$--$8$B). $y$-axes differ by identity
ceiling (Phi-3.5 $100\%$, Llama-3 $90\%$). Early-layer patches recover
the most failed examples.}
\label{fig:patching}
\end{figure*}

Figure~\ref{fig:patching} shows the single-layer patching recovery
curves for the two confirmed-gate models (\S\ref{sec:claim4}).

\begin{figure}[!htbp]
\centering
\includegraphics[width=\linewidth]{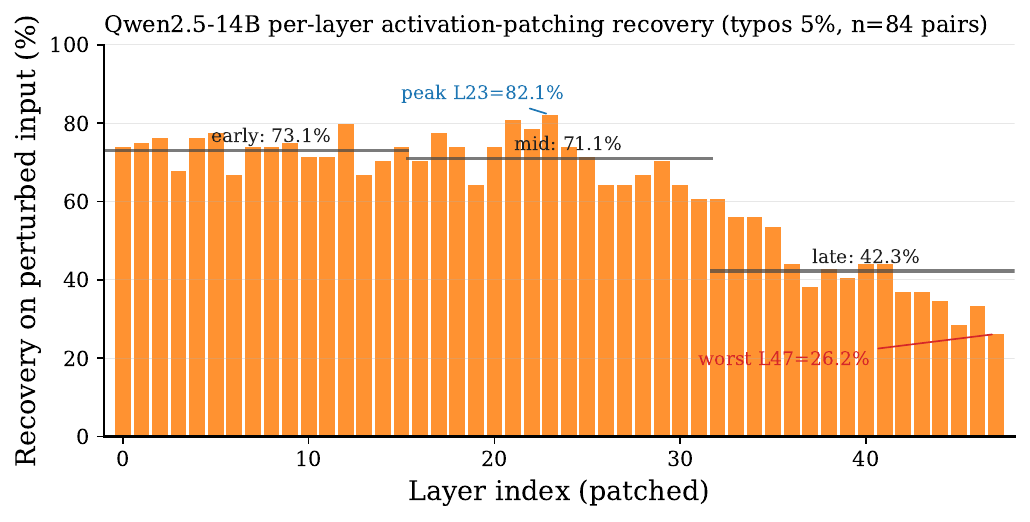}
\caption{Qwen2.5-14B activation-patching recovery across all $48$
layers on GSM8K typos~$5\%$. Identity recovery $85.0\%$ (above the
$80\%$ gate). The peak is at $L23$ ($82.1\%$); the worst single layer
is $L47$ ($26.2\%$). The random-patch control on this model recovered
$20.5\%$ (logged warning; \S\ref{sec:claim4}).}
\label{fig:supp_qwen14b_perlayer}
\end{figure}

Figure~\ref{fig:supp_qwen14b_perlayer} gives the full $48$-layer
activation-patching recovery curve for Qwen2.5-14B on GSM8K typos~$5\%$
($84$ clean-correct pairs of $100$ attempted). The main paper
(\S\ref{sec:claim4}) summarises the curve into per-region recovery and the per-layer peak
($L23$, $82.1\%$) and worst ($L47$, $26.2\%$); the figure shows that
the curve is broadly consistent with late-accumulation, with the
highest recovery concentrated in the early-to-mid third and a clear
collapse in the deepest layers.

\section{Per-Perturbation Recovery Rates}
\label{supp:recovery}
\label{sec:per_pert}

Table~\ref{tab:supp_recovery} gives per-perturbation recovery rates for all
five models (GSM8K). Cascade slope summaries are in
Table~\ref{tab:regime_summary}.

\begin{table*}[!htbp]
\centering
\small
\begin{tabular}{lccccc}
\toprule
Perturbation & Phi-3.5 & Gemma-2-9B & Llama-3 & Mistral & Qwen2.5-7B \\
\midrule
Typos 5\%       & 39.5\% & 34.5\% &  0.5\% &  1.5\% &  1.8\% \\
OCR 5\%         & 60.0\% & 35.5\% &  3.5\% &  4.5\% &  5.2\% \\
Whitespace 10\% & 43.5\% & 38.5\% &  0.0\% &  0.0\% &  0.8\% \\
Case 10\%       & 47.5\% & 36.5\% &  0.0\% &  0.0\% &  0.0\% \\
Homophones 20\% & 56.5\% & 51.5\% & 34.5\% & 38.0\% & 39.8\% \\
Speech 10\%     & 34.5\% & 21.0\% &  1.0\% &  0.5\% &  1.6\% \\
\bottomrule
\end{tabular}
\caption{Per-perturbation recovery rates for all five models (GSM8K).
Recovery is the fraction of layer-0 divergence eliminated by the final layer.
Spike-and-suppress models (Phi-3.5, Gemma-2-9B) show partial recovery on
all types; late-accumulation models (Llama-3, Mistral, Qwen2.5-7B) show
near-zero recovery, except homophones at 20\% severity (the only condition
producing meaningful late-model recovery).
Sample sizes: Phi-3.5, Mistral, Qwen2.5-7B $n=500$; Llama-3, Gemma-2-9B
$n=200$.}
\label{tab:supp_recovery}
\end{table*}

\section{Pairwise Spearman Correlations: All Five Models}
\label{supp:correlations}

Table~\ref{tab:correlations} reports the full pairwise correlation
panel underlying \S\ref{sec:threemap}: for each model, Spearman $\rho$
across layers between each pair of the three signals, with CIs and
$p$-values from the moving-block bootstrap (\S\ref{sec:stats_method}).
The pattern the main text summarizes is visible directly: LRD
$\times$ Patch is strongly negative on the confirmed-gate models
(Phi-3.5, Llama-3) and directionally negative on Mistral; LRD
$\times$ LoRA flips sign with regime but reaches per-model
significance only on Mistral (adapters from the same shared-grid lr
later diagnosed as lr-sensitive, \S\ref{sec:tally}; retained here as
directional evidence, consistent with Mistral's exclusion from sweep
adjudication); and the two models that fail the
identity-patch gate (Gemma-2-9B, Qwen2.5-7B) produce weak or unstable
correlations, consistent with a corrupted patching signal rather than
a contradiction.

\begin{table*}[!htbp]
\centering
\small
\begin{tabular}{llccc}
\toprule
Model & Pair & $\rho$ & 95\% CI & $p$ \\
\midrule
Phi-3.5    & LRD $\times$ Patch   & $-0.881$ & $[-0.991,\ -0.787]$ & ${<}0.001^{***}$ \\
           & LRD $\times$ LoRA    & $-0.550$ & $[-0.884,\ +0.624]$ & $0.33$ \\
           & Patch $\times$ LoRA  & $+0.482$ & $[-0.840,\ +0.909]$ & $0.44$ \\
\midrule
Llama-3    & LRD $\times$ Patch   & $-0.723$ & $[-0.982,\ -0.450]$ & ${<}0.001^{***}$ \\
           & LRD $\times$ LoRA    & $+0.631$ & $[-0.018,\ +0.908]$ & $0.054$ \\
           & Patch $\times$ LoRA  & $-0.642$ & $[-0.760,\ +0.091]$ & $0.072$ \\
\midrule
Mistral    & LRD $\times$ Patch   & $-0.752$ & $[-0.969,\ -0.550]$ & ${<}0.001^{***}$ \\
           & LRD $\times$ LoRA    & $+0.685$ & $[+0.132,\ +0.854]$ & $0.025^{*}$ \\
           & Patch $\times$ LoRA  & $-0.731$ & $[-0.824,\ -0.080]$ & $0.033^{*}$ \\
\midrule
Gemma-2-9B & LRD $\times$ Patch   & $+0.389$ & $[-0.43,\ +0.83]$ & $0.264$ \\
           & LRD $\times$ LoRA    & $+0.091$ & $[-0.38,\ +0.69]$ & $0.418$ \\
           & Patch $\times$ LoRA  & $+0.735$ & $[+0.26,\ +0.92]$ & $0.006^{**\dag}$ \\
\midrule
Qwen2.5-7B & LRD $\times$ Patch   & $+0.025$ & $[-0.575,\ +0.728]$ & $0.866$ \\
           & LRD $\times$ LoRA    & $+0.413$ & $[-0.763,\ +0.726]$ & $0.802$ \\
           & Patch $\times$ LoRA  & $-0.167$ & $[-0.529,\ +0.356]$ & $0.602$ \\
\bottomrule
\end{tabular}
\caption{Pairwise Spearman correlations (block bootstrap, block size~5, 10k
resamples). $^{*}p<0.05$, $^{**}p<0.01$, $^{***}p<0.001$.
$\dag$~Gemma-2-9B patching is exploratory (60\% identity ceiling).
LRD $\times$ LoRA is directionally consistent with regime but significant only
on Mistral after autocorrelation correction; Mistral's LoRA data here is
from the same shared-grid lr later diagnosed as lr-sensitive
(\S\ref{sec:tally}) and is retained as directional/exploratory evidence only.}
\label{tab:correlations}
\end{table*}

\begin{figure*}[t]
\centering
\includegraphics[width=\linewidth]{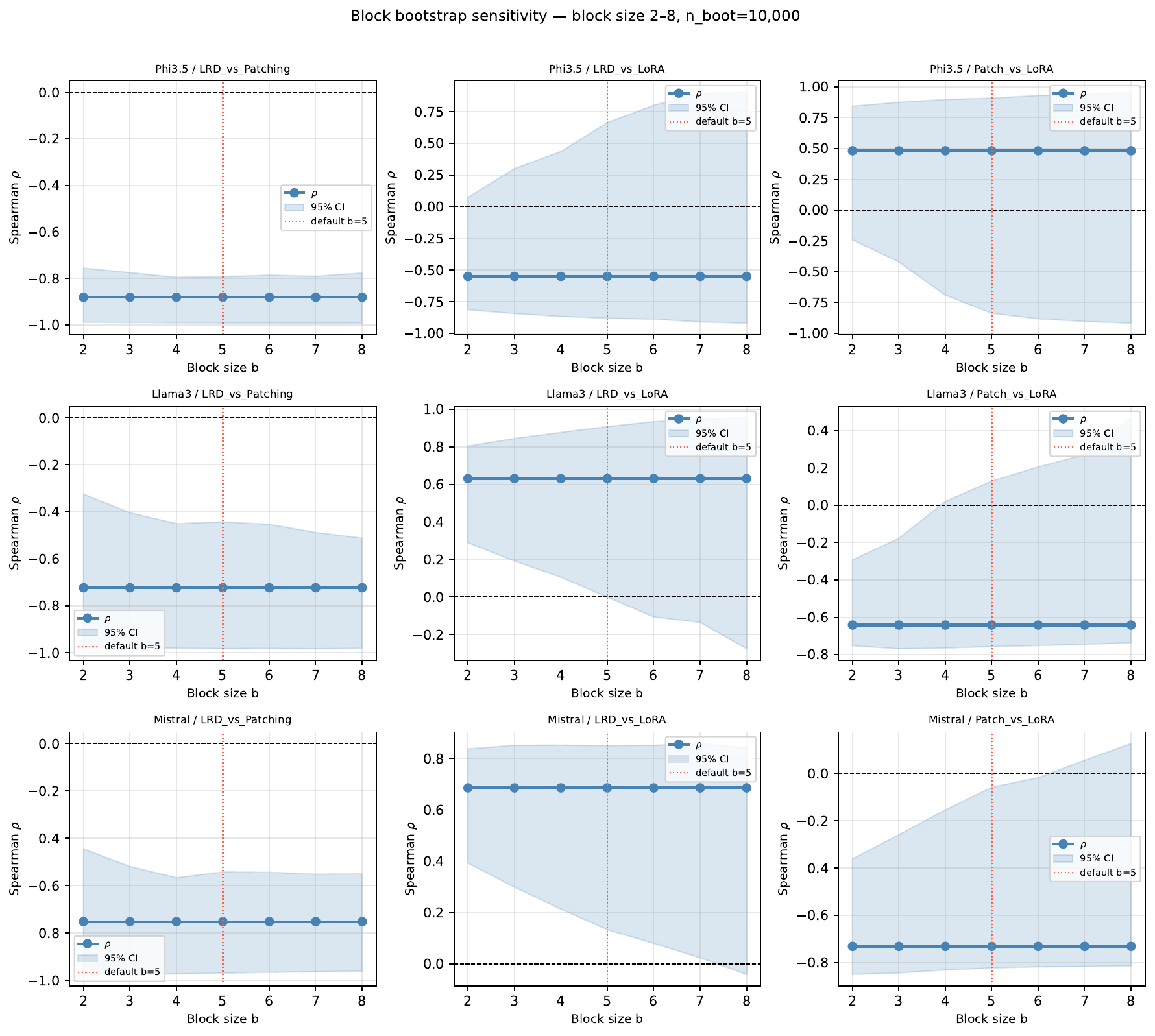}
\caption{Block-size sensitivity for the moving-block bootstrap
(Table~\ref{tab:correlations}), $b{=}2$--$8$, $10$k resamples per $b$.
Point estimate $\rho$ is a property of the fitted layer-pair (not the
resample), so it is constant across $b$; the shaded band is the 95\% CI.
The LRD$\times$Patch anti-correlation clears $p<0.05$ at every $b$ for
all three models (dashed red line marks the reported default $b{=}5$);
LRD$\times$LoRA and Patch$\times$LoRA lose significance at some block
sizes, matching the sign-not-magnitude claim in
\S\ref{sec:stats_method}.}
\label{sec:block_sensitivity}
\end{figure*}

\section{Intrinsic Predictor Details}
\label{supp:intrinsic}
\label{sec:intrinsic}

We evaluated four intrinsic metrics as candidate predictors of
per-layer LoRA effectiveness. C3 and C4 are defined in
\S\ref{sec:stats_method} of the main paper; we repeat them here with
implementation detail, alongside C1 and C2.

\begin{itemize}
\item \textbf{C1: Update ratio.}
  $C_1(L) = \|\mathbf{h}^{(L)} - \mathbf{h}^{(L-1)}\|_2 \,/\, \|\mathbf{h}^{(L-1)}\|_2$,
  where $\mathbf{h}^{(L)}$ is the hidden state at layer~$L$, mean-pooled over
  valid (non-padded) token positions per example. Computed on $n=200$ clean
  GSM8K examples and averaged across examples.
\item \textbf{C2: Attention entropy.}
  $C_2(L) = -\sum_{k} p_k \log p_k$, where $p_k$ are the attention
  probabilities at layer~$L$ (extracted via \texttt{output\_attentions=True}).
  Entropy is computed per-head per-query-position, then averaged across
  (batch, heads, query positions) on $n=200$ clean GSM8K examples.
\item \textbf{C3: Weight effective rank (stable rank).}
  $C_3(L) = \exp\!\left(-\sum_i \tilde{\sigma}_i \log \tilde{\sigma}_i\right)$,
  with $\tilde{\sigma}_i = \sigma_i / \sum_j \sigma_j$ and $\sigma_i$ the
  singular values of the layer's weight matrix. This is a smooth, differentiable
  rank proxy (not a threshold-based count). Computed statically on
  \texttt{q\_proj} and \texttt{v\_proj}; for Phi-3.5 fused
  \texttt{qkv\_proj}, the Q and V slices are extracted and the per-layer
  score is $(r_Q + r_V)/2$. No forward pass required.
\item \textbf{C4: Gradient norm.}
  $C_4(L) = \tfrac{1}{2}\!\left(
    \big\|\partial \mathcal{L}_{\text{CE}} / \partial W_Q^{(L)}\big\|_F
    + \big\|\partial \mathcal{L}_{\text{CE}} / \partial W_V^{(L)}\big\|_F
  \right)$,
  the mean of the Frobenius gradient norms of the Q and V weight matrices
  under teacher-forced cross-entropy loss. Computed on $n=50$ clean GSM8K
  examples (subset of the 200) with batch size~2; requires forward+backward
  pass.
\end{itemize}

C3 and C4 show the strongest \emph{post-hoc} correlations with LoRA
effectiveness on the original three models
(Table~\ref{tab:intrinsic}); C1 and C2 are reported in the same table
for completeness. As reported in the main paper (\S\ref{sec:threemap}),
the prospective validation on Qwen-7B (C3/C4 predicted L08--11) is
falsified by the fixed-harness layer sweep: L08--11 is the worst
window for Qwen-7B ($-7.8$~pp mean perturbed $\Delta$, three seeds),
not the best. We do not recommend C3/C4 as a placement rule.

\begin{table}[H]
\centering
\small
\resizebox{\columnwidth}{!}{\begin{tabular}{lrrr}
\toprule
Metric & Phi-3.5 $\rho$ & Llama-3 $\rho$ & Mistral $\rho$ \\
\midrule
C1: Update ratio      & $-0.517^{**}$  & $+0.394^{*}$   & $+0.430^{*}$   \\
C2: Attn entropy      & $+0.211$        & $-0.712^{***}$ & $-0.296$        \\
C3: Weight eff.\ rank & $-0.627^{***}$ & $+0.751^{***}$ & $+0.670^{***}$ \\
C4: Gradient norm     & $+0.537^{**}$  & $-0.714^{***}$ & $-0.594^{***}$ \\
\bottomrule
\end{tabular}}
\caption{Intrinsic capacity metric correlations with LoRA effectiveness
(original three models; i.i.d.\ bootstrap, anti-conservative given
eff~$N\approx3$--5; significance markers are indicative). C3 and C4 flip sign
between Phi-3.5 and the late-accumulation models.}
\label{tab:intrinsic}
\end{table}

Table~\ref{tab:supp_c3c4} summarises C3/C4 predictions versus the
least-regressive window under the fixed-harness layer sweep
(\S\ref{sec:sweep}). The original-three-model fits stay
post-hoc; on the two pre-registered models, the prediction is
falsified on Qwen-7B and is moot on Gemma (flat sweep).
\label{supp:c3c4}

\begin{table}[H]
\centering
\small
\resizebox{\columnwidth}{!}{\begin{tabular}{lccc}
\toprule
Model      & C3/C4 predicted & Least-regr.\ window & Outcome \\
\midrule
Phi-3.5    & L15--19 & L27--31 & Post-hoc only; sign flipped \\
Llama-3    & L20--24 & L27--31 & Post-hoc only; close, not best \\
Mistral    & L20--24 & L27--31 & Post-hoc only; close, not best \\
Qwen2.5-7B & L08--11 & L24--27 & \textbf{Falsified} (predicted = worst) \\
Gemma-2-9B & L24--29 & (flat)  & Moot ($\rho \approx 0$) \\
\bottomrule
\end{tabular}}
\caption{C3/C4 predicted window versus the fixed-harness
least-regressive window. The Phi-3.5/Llama-3/Mistral entries are
post-hoc fits to the broken-harness sweep; under the fixed harness no
window is non-negative on Phi-3.5 or Mistral, and the least-regressive
window is L27--31 on all three. On Qwen-7B, the C3/C4 prediction is the
\emph{worst} window in the sweep ($-7.8$~pp; pre-registered falsifier).
Gemma's sweep is flat.}
\label{tab:supp_c3c4}
\end{table}

\section{Three-Map Overlays: All Five Models}
\label{sec:three_map_overlays}

\begin{figure*}[!htbp]
\centering
\begin{subfigure}{0.49\textwidth}
  \centering
  \includegraphics[width=\linewidth]{figures/three_map_phi35.pdf}
  \subcaption{Phi-3.5 (S\&S): LoRA peaks where LRD is suppressed.}
  \label{fig:supp_three_map_phi}
\end{subfigure}\hfill
\begin{subfigure}{0.49\textwidth}
  \centering
  \includegraphics[width=\linewidth]{figures/three_map_llama3.pdf}
  \subcaption{Llama-3 (Late): LoRA peaks where LRD is building.}
  \label{fig:supp_three_map_llama}
\end{subfigure}
\\[0.4em]
\begin{subfigure}{0.49\textwidth}
  \centering
  \includegraphics[width=\linewidth]{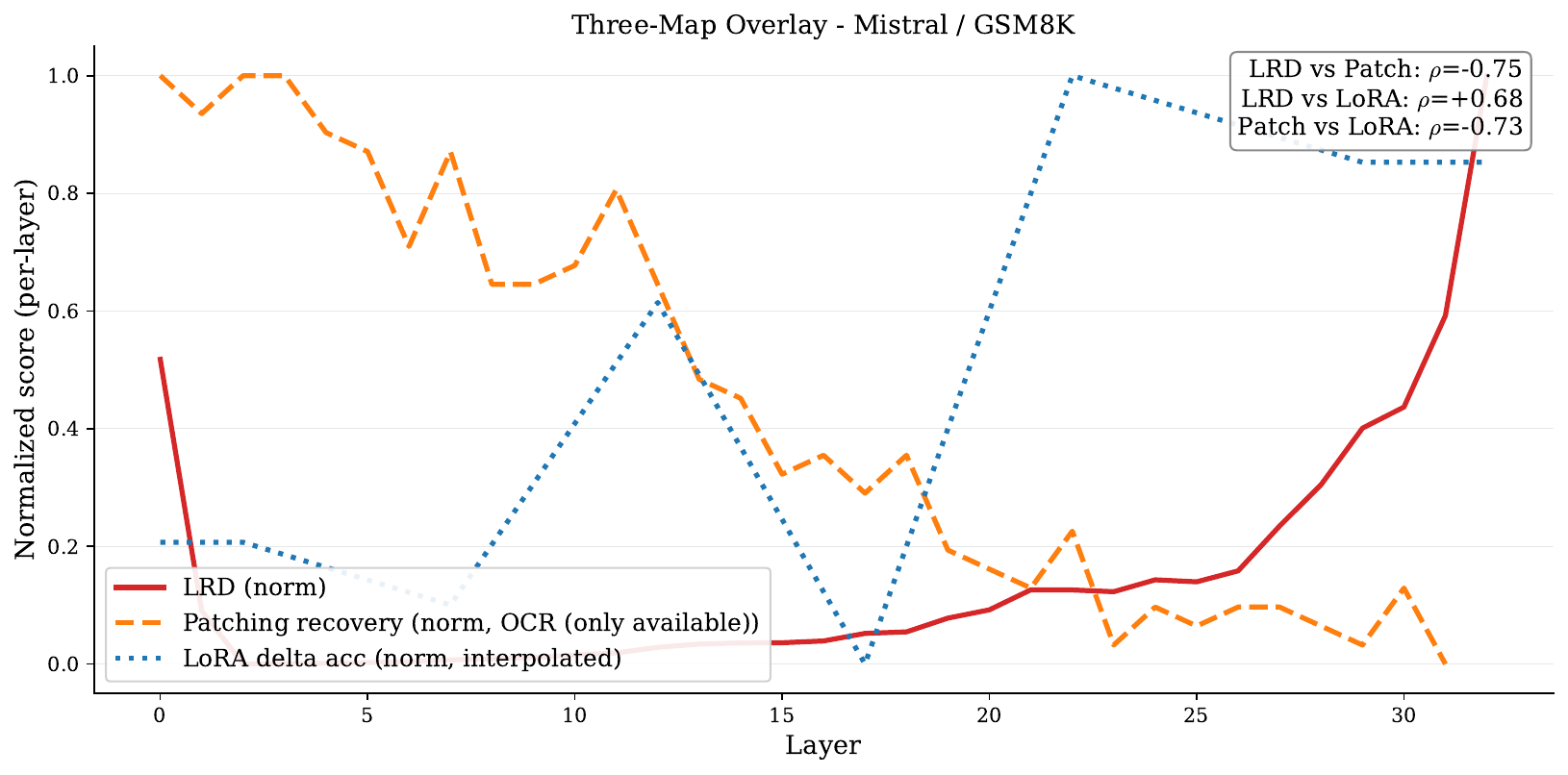}
  \subcaption{Mistral (Late): LoRA anti-correlated with patching window.}
  \label{fig:supp_three_map_mistral}
\end{subfigure}\hfill
\begin{subfigure}{0.49\textwidth}
  \centering
  \includegraphics[width=\linewidth]{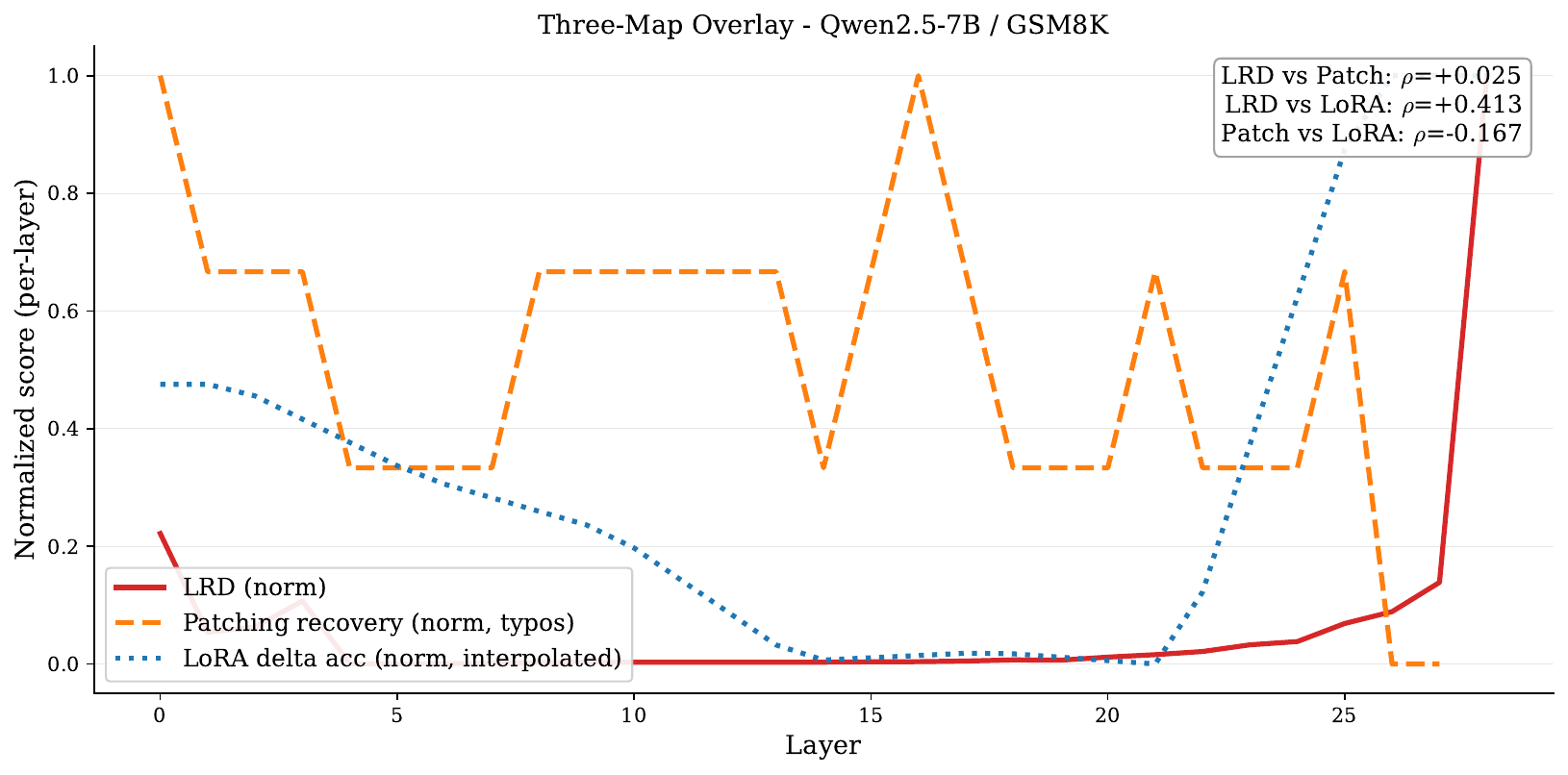}
  \subcaption{Qwen2.5-7B (Late): LRD$\times$Patch unusable ($n{=}3$).}
  \label{fig:supp_three_map_qwen}
\end{subfigure}
\\[0.4em]
\begin{subfigure}{0.49\textwidth}
  \centering
  \includegraphics[width=\linewidth]{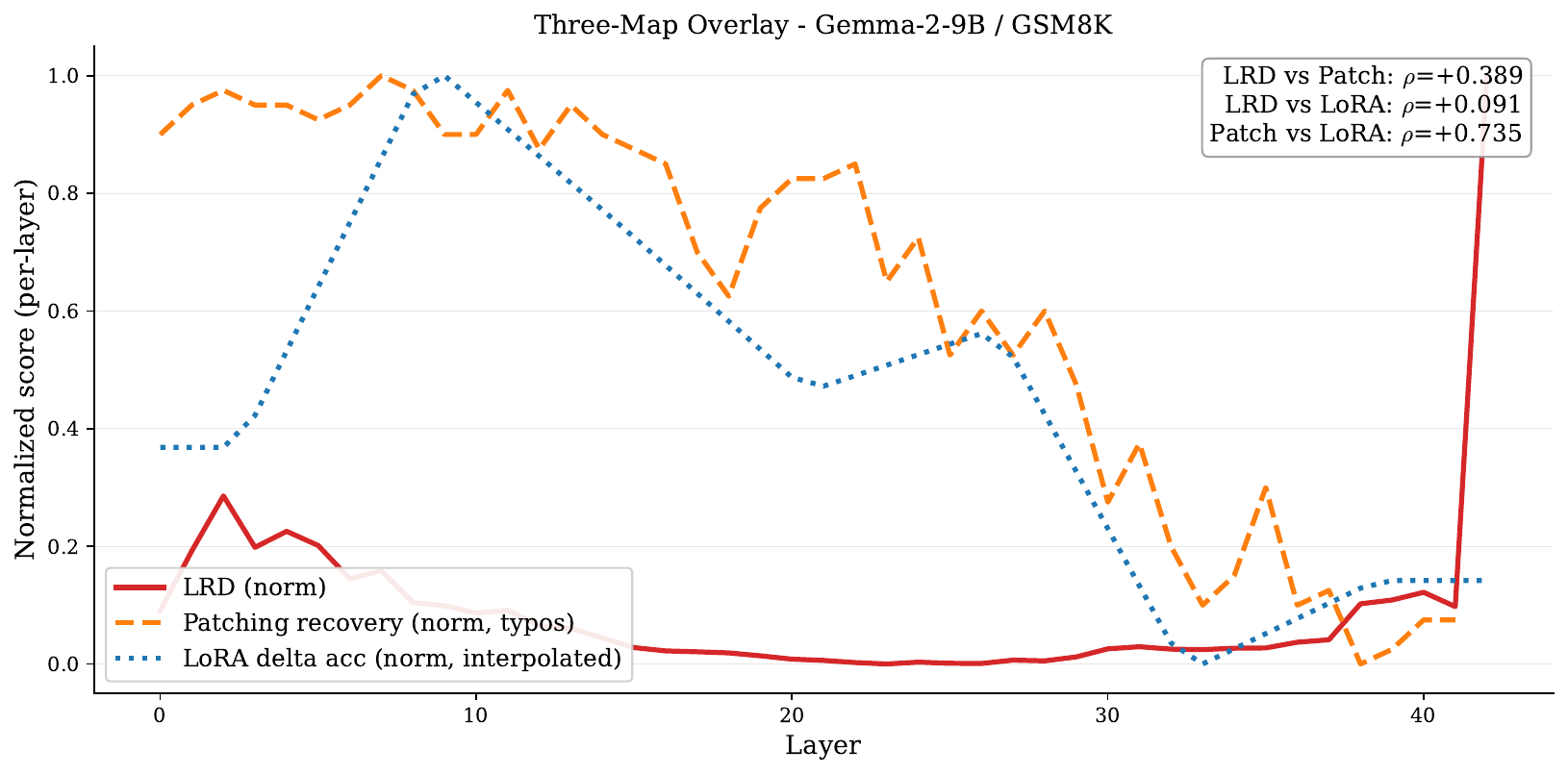}
  \subcaption{Gemma-2-9B (S\&S): exception, $\rho{=}{+}0.389$ ($60\%$ ceiling).}
  \label{fig:supp_three_map_gemma}
\end{subfigure}
\caption{\textbf{Three-map overlays, all five models.} LRD sensitivity,
patching recovery, and LoRA effectiveness per layer.}
\label{fig:supp_three_map}
\end{figure*}

Figure~\ref{fig:supp_three_map} shows the per-layer overlay of the three
signals (LRD sensitivity, patching recovery, LoRA effectiveness) for
each of the five models in the panel; the Phi-3.5 and Llama-3 panels
also appear as the anchor figure in the main paper
(Figure~\ref{fig:threemap_main}). Numerical correlations across
the panels are in Table~\ref{tab:correlations}.

\section{Per-Layer Cohen's $d$: All Five Models}
\label{sec:cohens_d_panels}

\begin{figure*}[!htbp]
\centering
\begin{subfigure}{0.49\textwidth}
  \centering
  \includegraphics[width=\linewidth]{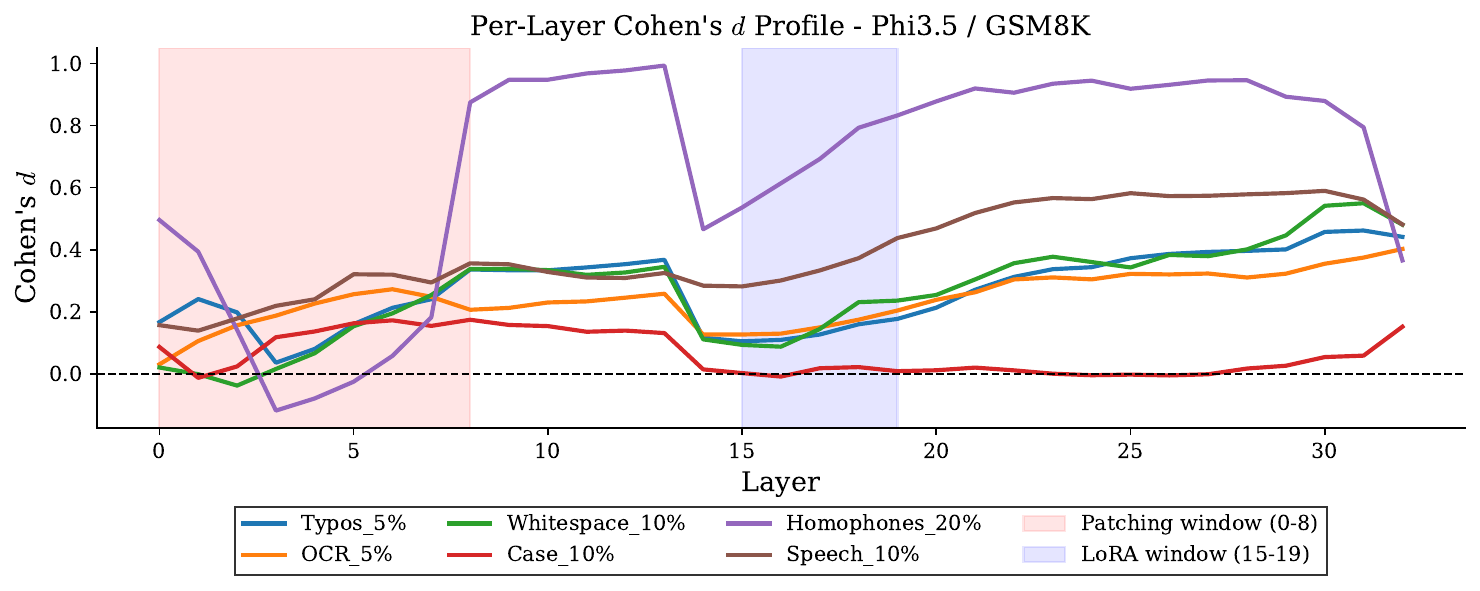}
  \subcaption{Phi-3.5 (S\&S): peak at L28--32.}
  \label{fig:supp_cohens_d_phi}
\end{subfigure}\hfill
\begin{subfigure}{0.49\textwidth}
  \centering
  \includegraphics[width=\linewidth]{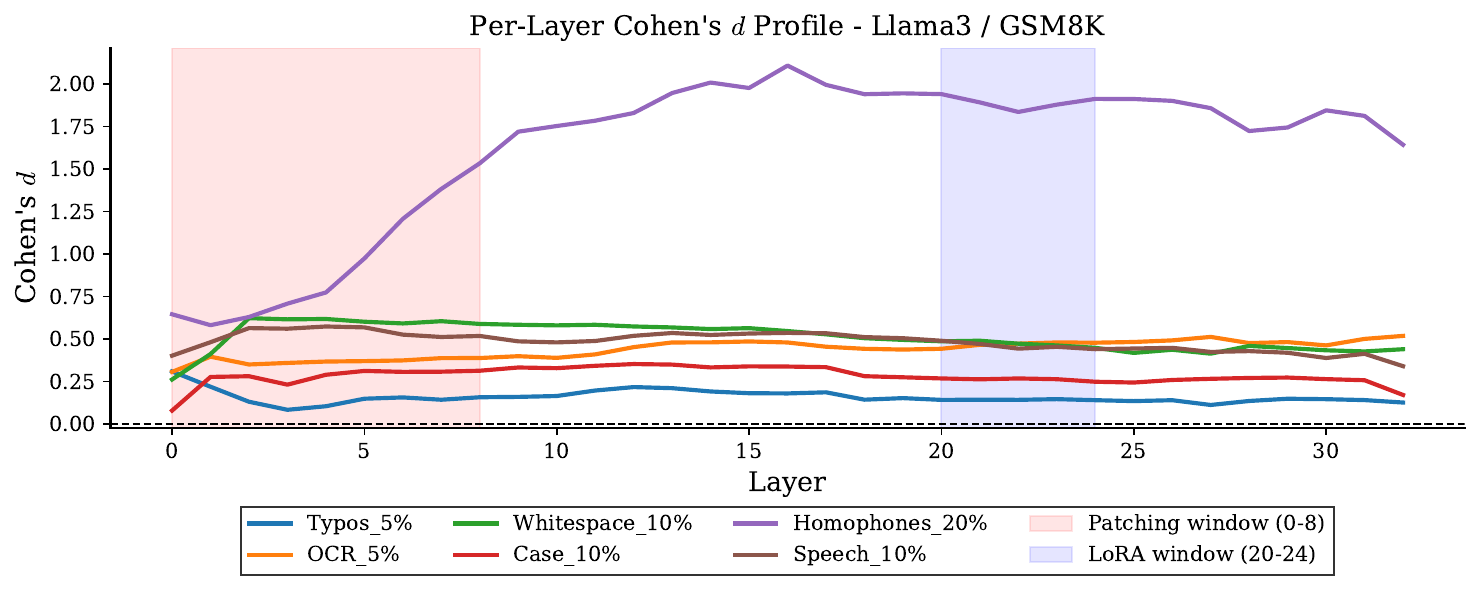}
  \subcaption{Llama-3 (Late).}
  \label{fig:supp_cohens_d_llama}
\end{subfigure}
\\[0.4em]
\begin{subfigure}{0.49\textwidth}
  \centering
  \includegraphics[width=\linewidth]{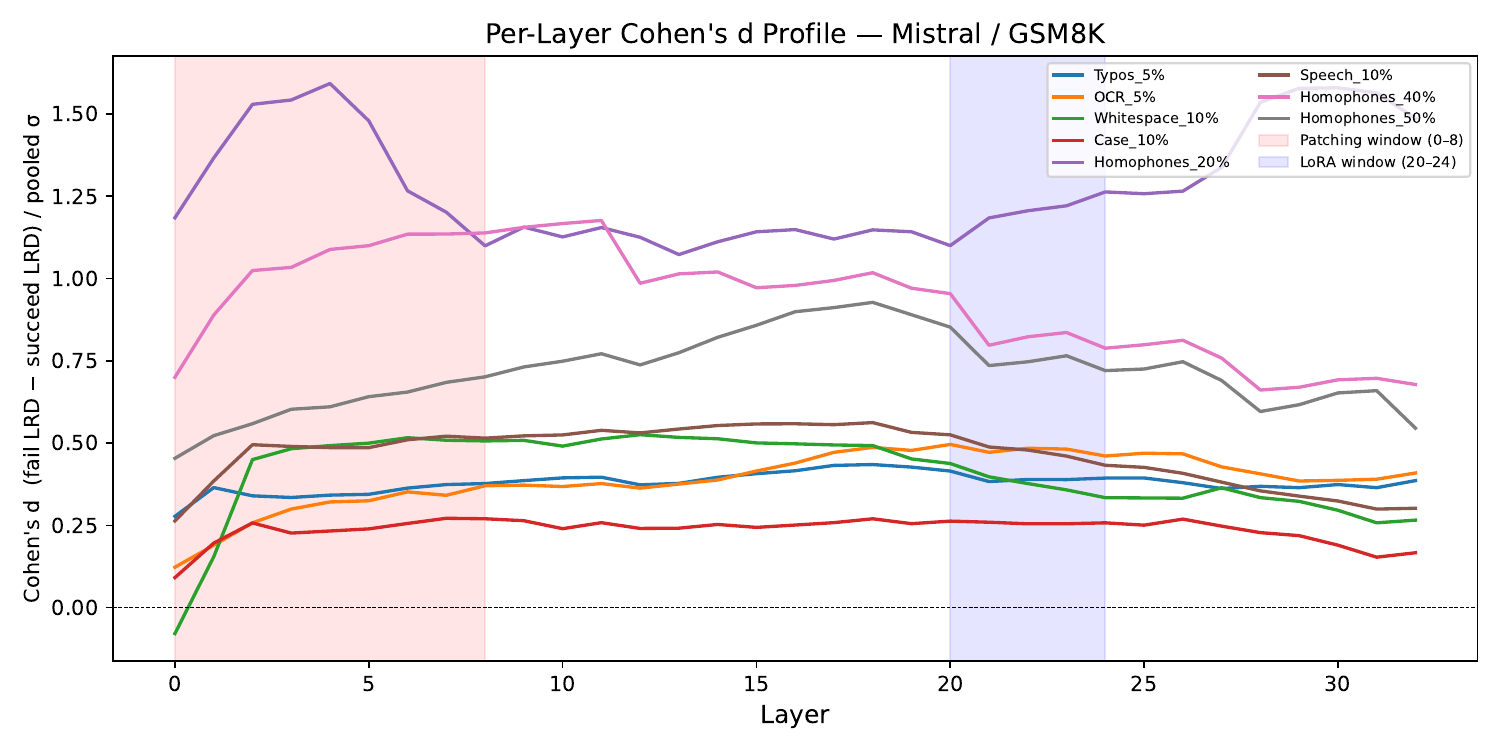}
  \subcaption{Mistral (Late).}
  \label{fig:supp_cohens_d_mistral}
\end{subfigure}\hfill
\begin{subfigure}{0.49\textwidth}
  \centering
  \includegraphics[width=\linewidth]{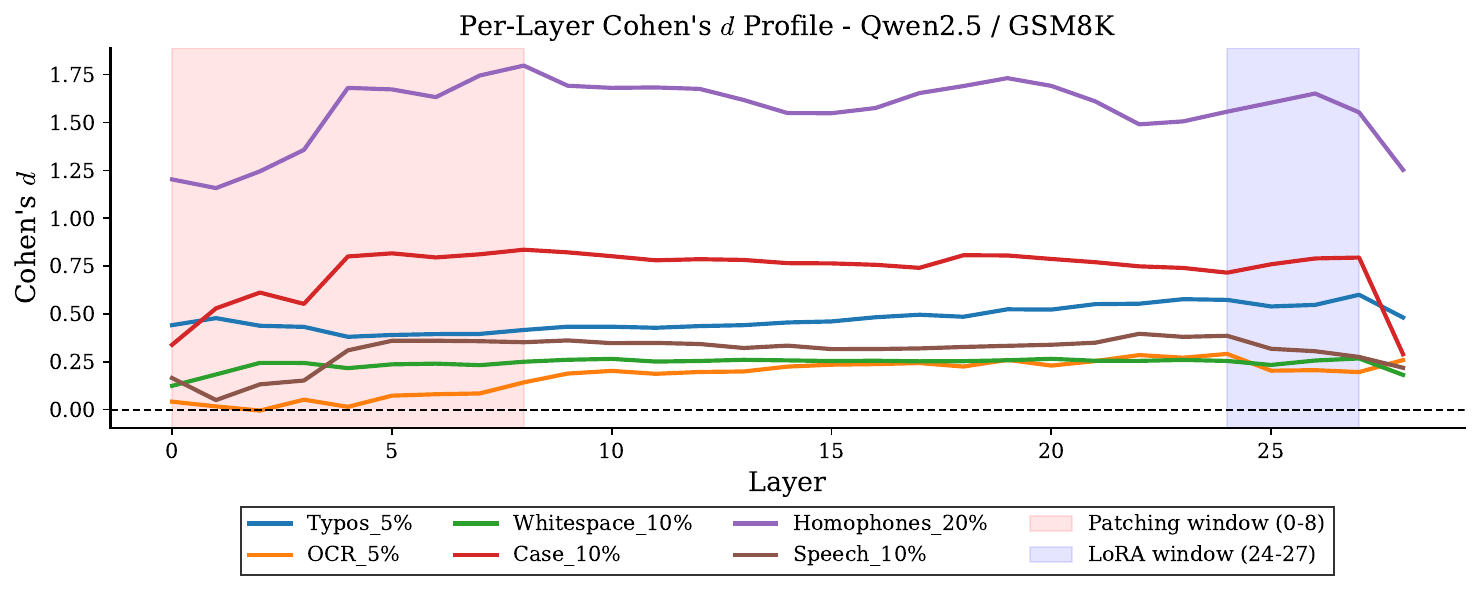}
  \subcaption{Qwen2.5-7B (Late): peak $d{\approx}1.80$ at L8 (Homophones~20\%).}
  \label{fig:supp_cohens_d_qwen}
\end{subfigure}
\\[0.4em]
\begin{subfigure}{0.49\textwidth}
  \centering
  \includegraphics[width=\linewidth]{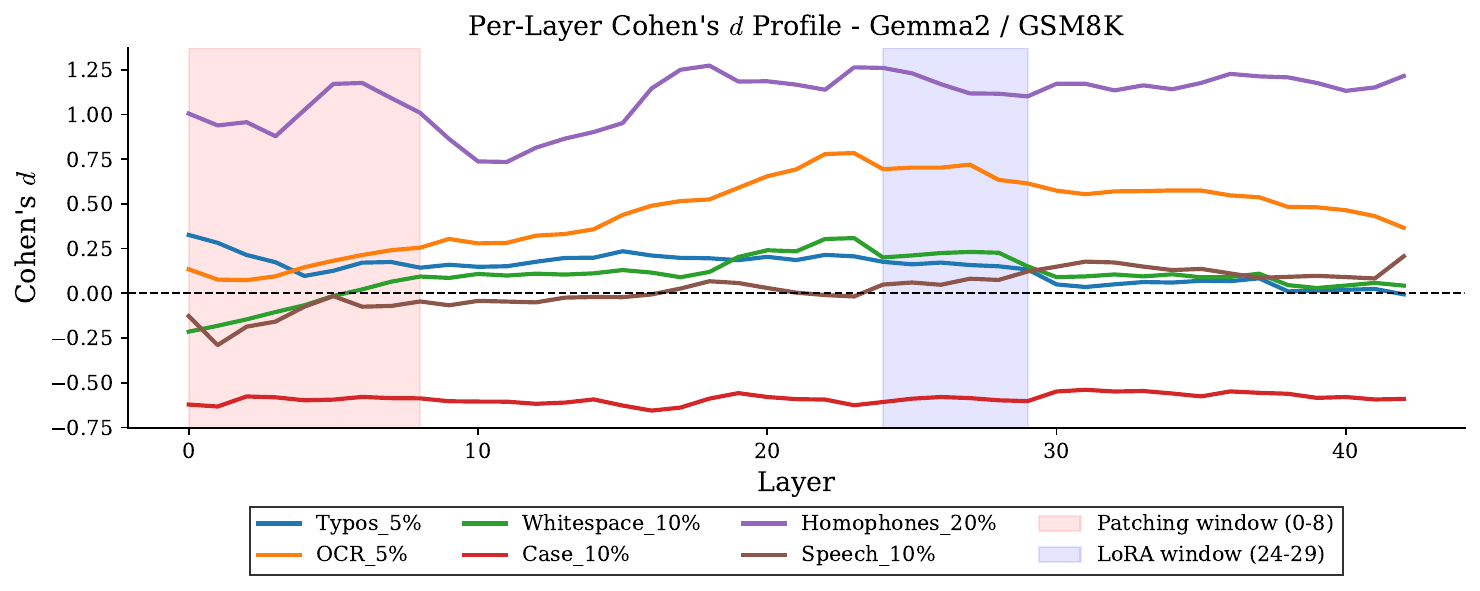}
  \subcaption{Gemma-2-9B (S\&S): peak $d{\approx}1.27$ at L18 (Homophones~20\%).}
  \label{fig:supp_cohens_d_gemma}
\end{subfigure}
\caption{\textbf{Per-layer Cohen's $d$, all five models.} Shaded band:
activation-patching causal window. Effect sizes peak in the deep layers,
downstream of every adapter window tested under the fixed harness.}
\label{fig:supp_cohens_d}
\end{figure*}

Figure~\ref{fig:supp_cohens_d} shows the per-layer Cohen's $d$ on
failing vs.\ succeeding examples for each model
(\S\ref{sec:claim3}). Shaded bands mark each model's
activation-patching causal window.
Effect sizes peak in the deep layers, downstream of every adapter
window tested in the fixed-harness sweep (\S\ref{sec:sweep}).

\section{Intrinsic Capacity Metrics: Llama-3 and Mistral}
\label{sec:capacity_panels}

\begin{figure*}[!htbp]
\centering
\begin{subfigure}{\textwidth}
  \centering
  \includegraphics[width=0.9\textwidth]{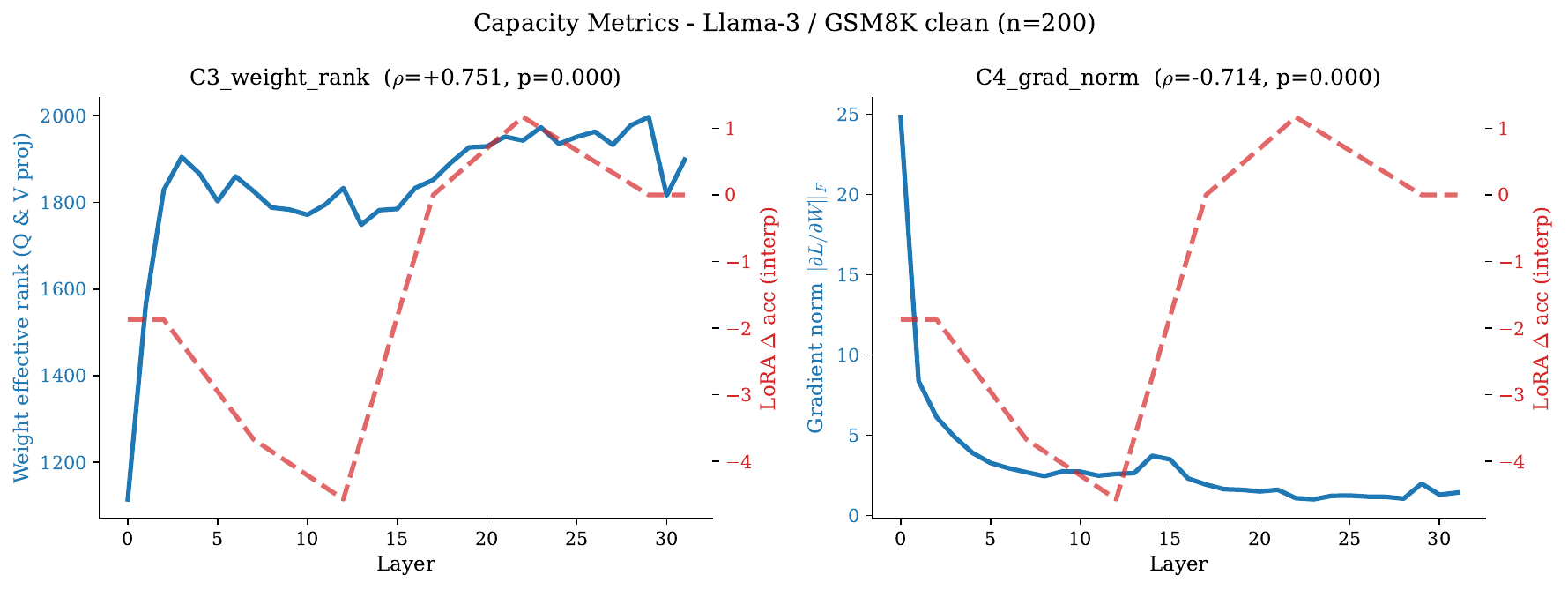}
  \caption{Llama-3.}
  \label{fig:supp_capacity_llama}
\end{subfigure}
\\[0.5em]
\begin{subfigure}{\textwidth}
  \centering
  \includegraphics[width=0.9\textwidth]{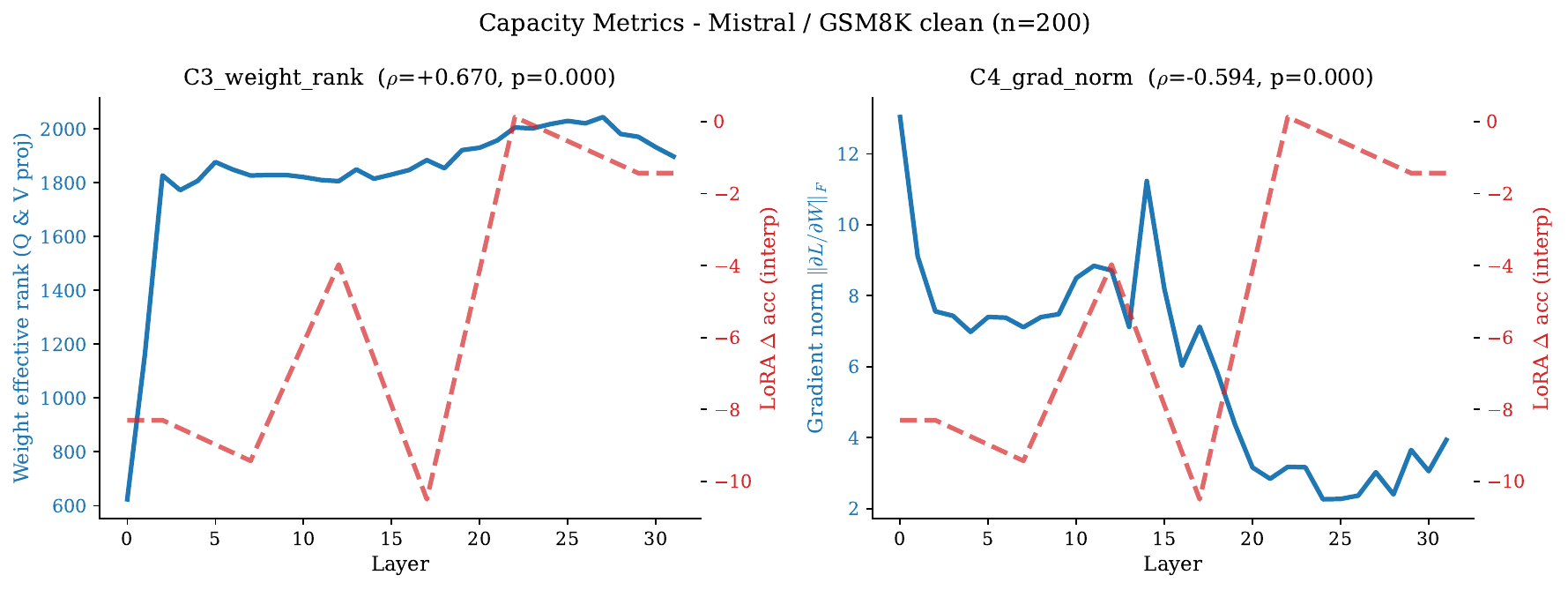}
  \caption{Mistral.}
  \label{fig:supp_capacity_mistral}
\end{subfigure}
\caption{\textbf{Intrinsic Capacity Metrics: Llama-3 and Mistral.} C3 (weight
effective rank, left) and C4 (gradient norm, right) for each model. Both
models show the opposite sign pattern from Phi-3.5 (correlations in
Table~\ref{tab:intrinsic}). C1 and C2 correlations are reported in
Table~\ref{tab:intrinsic}.}
\label{fig:supp_capacity}
\end{figure*}

Figure~\ref{fig:supp_capacity} shows C3 (weight effective rank) and C4
(gradient norm) for Llama-3 and Mistral. Both models show the opposite
sign pattern from Phi-3.5 (numerical correlations in
Table~\ref{tab:intrinsic}). These curves are descriptive: C3/C4 was
demoted to post-hoc only after the prospective validation on Qwen-7B
was falsified by the fixed-harness layer sweep (\S\ref{sec:threemap},
Table~\ref{tab:supp_c3c4}).

\section{Cascade Disruption: Remaining Panel Models}
\label{sec:cascade_per_model}

\begin{figure}[!htbp]
\centering
\includegraphics[width=\linewidth]{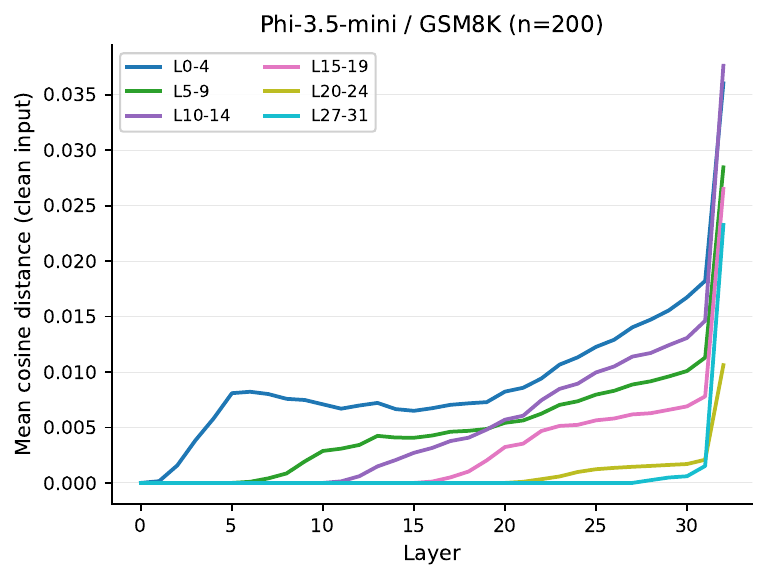}
\caption{Clean-input disruption for Phi-3.5 (one line per LoRA
window; $n{=}200$). Upstream of each window, divergence is exactly
zero; disruption onsets at the window and persists downstream,
largest for the earliest windows.}
\label{fig:disruption}
\end{figure}

\begin{figure*}[!htbp]
\centering
\begin{subfigure}{\textwidth}
  \centering
  \includegraphics[width=0.9\textwidth]{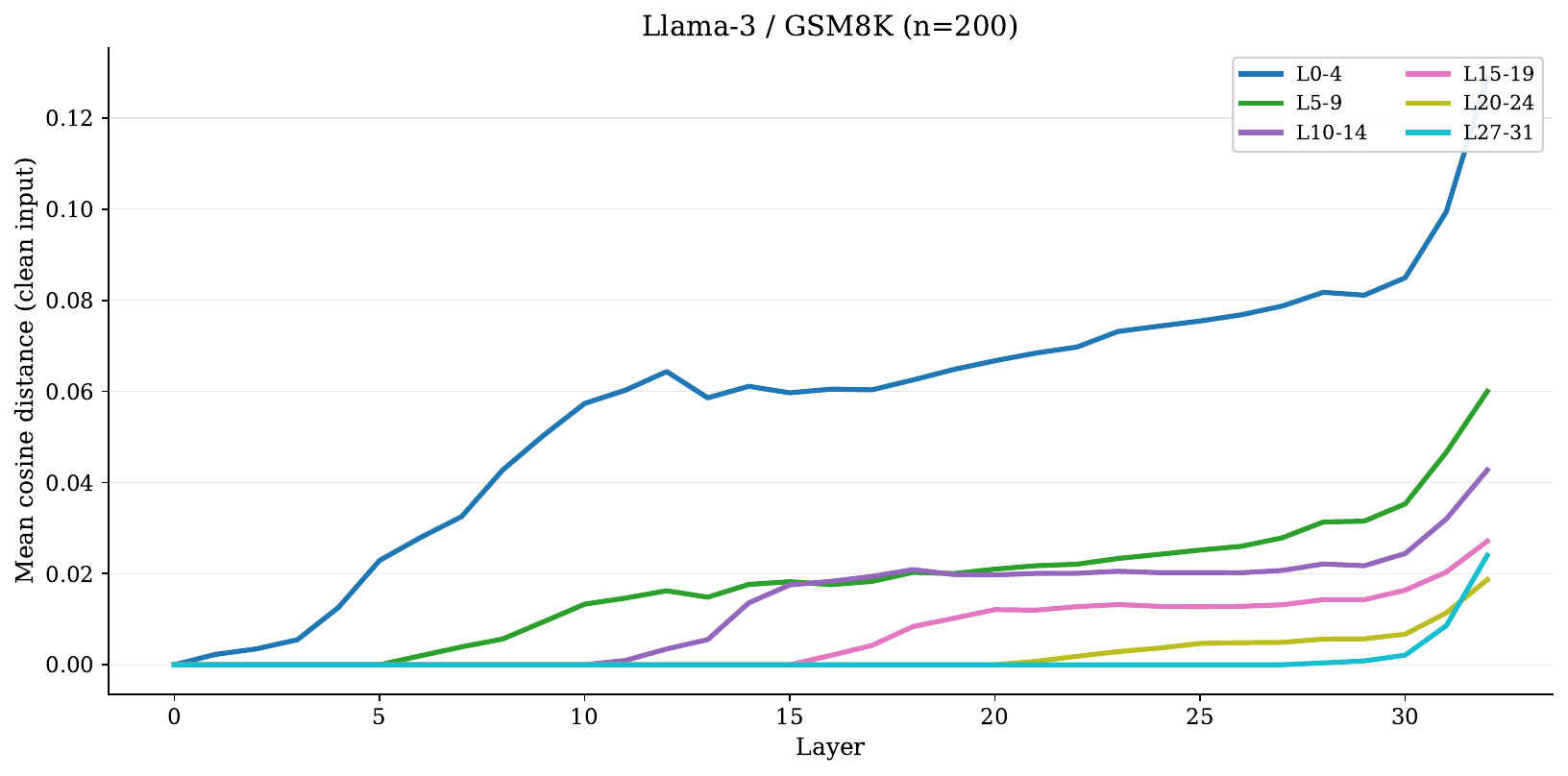}
  \caption{Llama-3.}
  \label{fig:supp_disruption_llama}
\end{subfigure}
\\[0.5em]
\begin{subfigure}{\textwidth}
  \centering
  \includegraphics[width=0.9\textwidth]{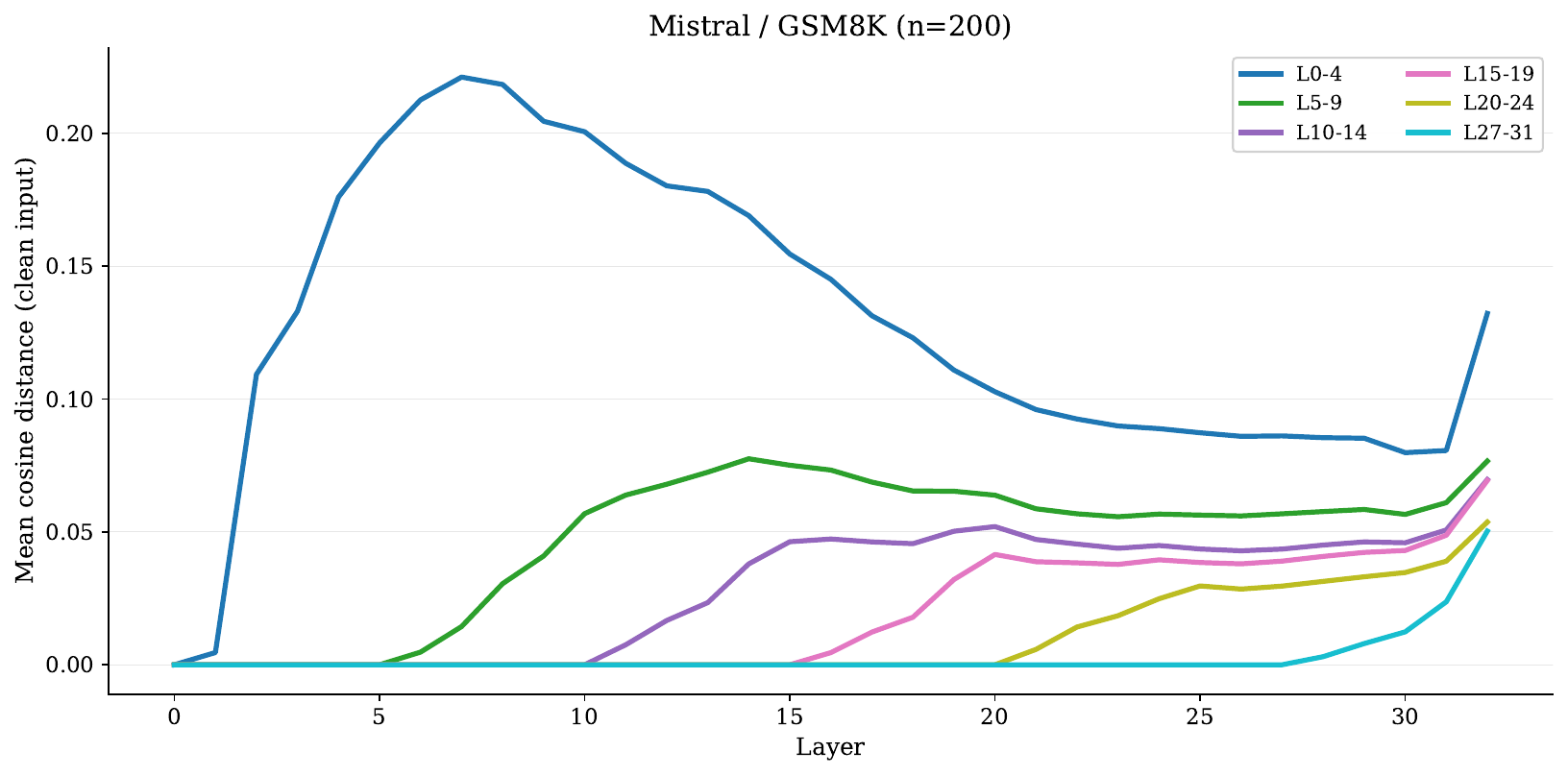}
  \caption{Mistral.}
  \label{fig:supp_disruption_mistral}
\end{subfigure}
\\[0.5em]
\begin{subfigure}{0.48\textwidth}
  \centering
  \includegraphics[width=\linewidth]{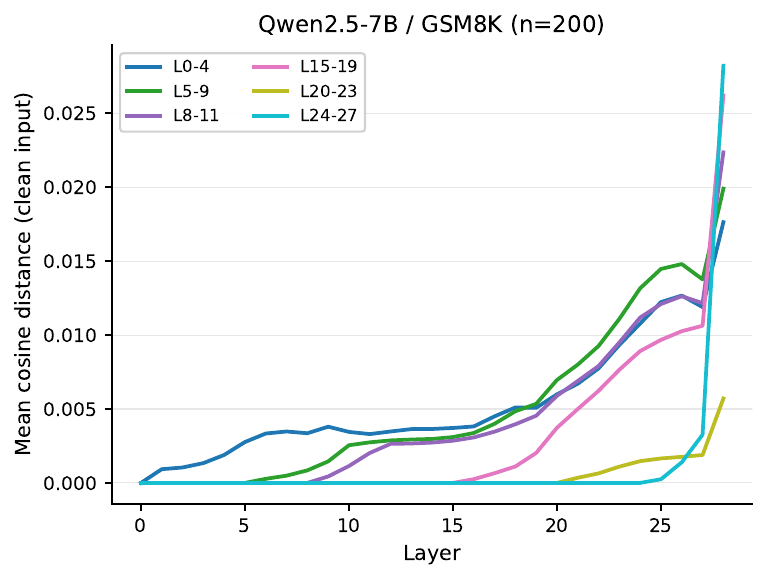}
  \caption{Qwen2.5-7B.}
  \label{fig:supp_disruption_qwen}
\end{subfigure}
\caption{\textbf{Cascade Disruption: Remaining Panel Models.}
Clean-input disruption (one line per LoRA window; $n{=}200$;
independently retrained adapters, cf.\
Appendix~\ref{sec:cascade_extra}). On all three models, divergence
is exactly zero upstream of the adapted window, onsets at the window,
and persists through every downstream layer; total disruption is
largest for the earliest windows and shrinks with window depth,
consistent with Phi-3.5 (Figure~\ref{fig:disruption}).}
\label{fig:supp_disruption}
\end{figure*}

Figure~\ref{fig:disruption} shows the Phi-3.5 cascade disruption
profiles referenced in \S\ref{sec:disruption};
Figure~\ref{fig:supp_disruption} shows Llama-3, Mistral, and
Qwen2.5-7B. All profiles are measured teacher-forced on shared clean
inputs, using the independently retrained sweep adapters of
Appendix~\ref{sec:cascade_extra}. Gemma-2-9B is omitted
(adapter-insensitive, \S\ref{sec:tally}; its original adapters were
unavailable for re-measurement).

\section{Clean Disruption vs.\ Perturbed $\Delta$: Independent Replication}
\label{supp:disruption_scatter}
\label{sec:cascade_extra}

\begin{figure*}[!htbp]
\centering
\begin{subfigure}{0.42\textwidth}
  \centering
  \includegraphics[width=\linewidth]{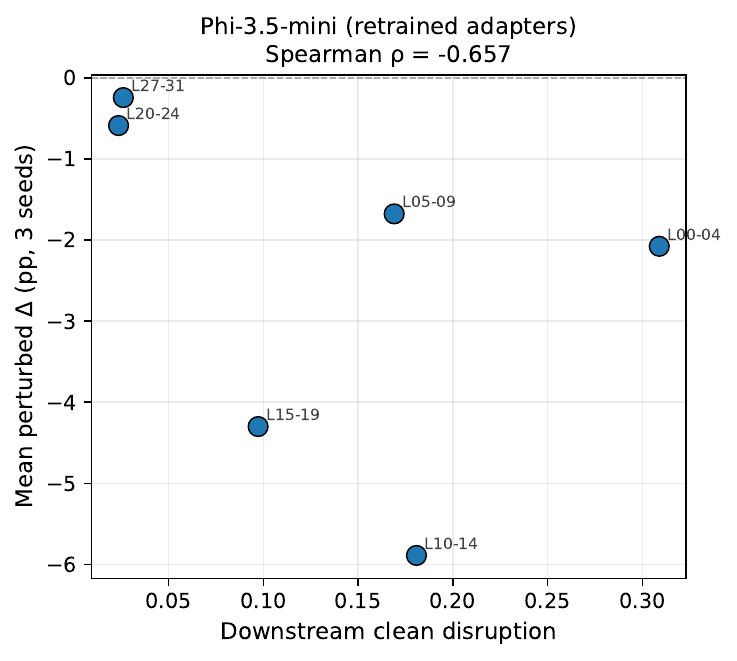}
  \subcaption{Phi-3.5.}
  \label{fig:supp_disruption_scatter_phi}
\end{subfigure}\hfill
\begin{subfigure}{0.42\textwidth}
  \centering
  \includegraphics[width=\linewidth]{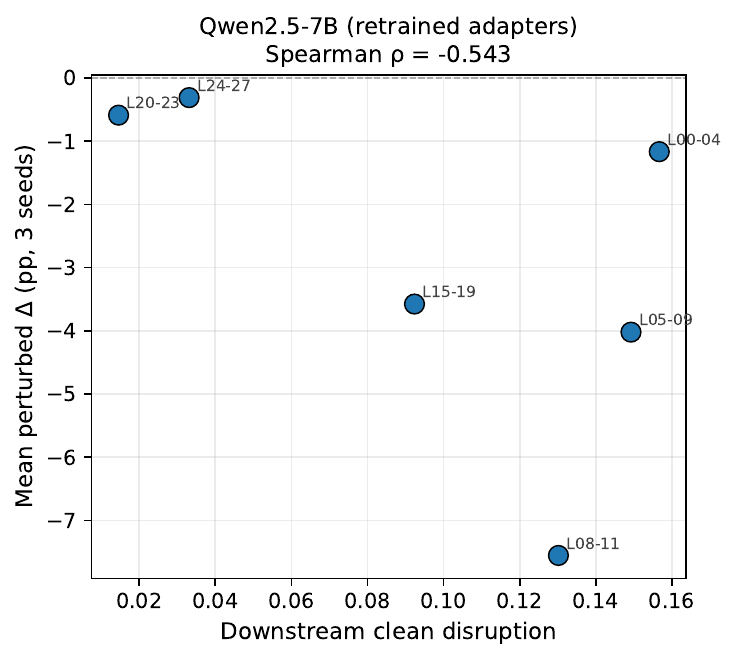}
  \subcaption{Qwen2.5-7B.}
  \label{fig:supp_disruption_scatter_qwen}
\end{subfigure}

\begin{subfigure}{0.42\textwidth}
  \centering
  \includegraphics[width=\linewidth]{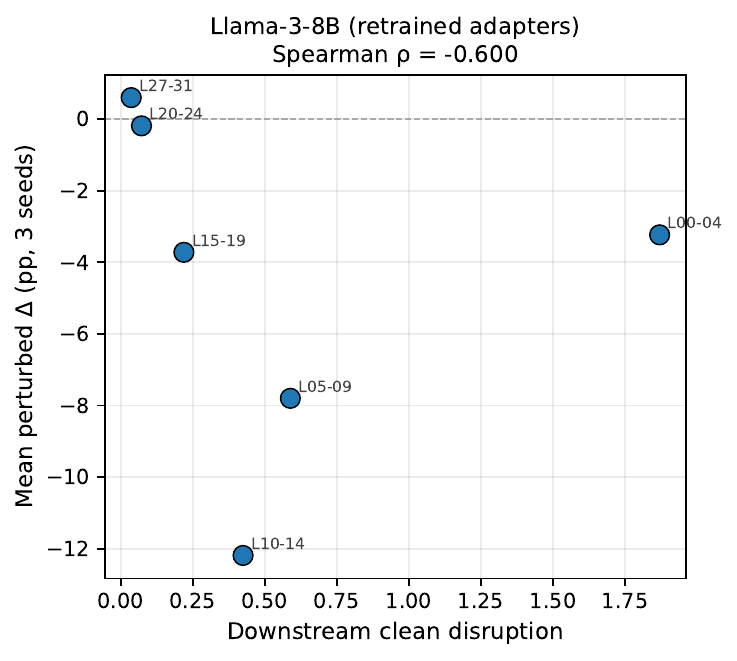}
  \subcaption{Llama-3.}
  \label{fig:supp_disruption_scatter_llama}
\end{subfigure}\hfill
\begin{subfigure}{0.42\textwidth}
  \centering
  \includegraphics[width=\linewidth]{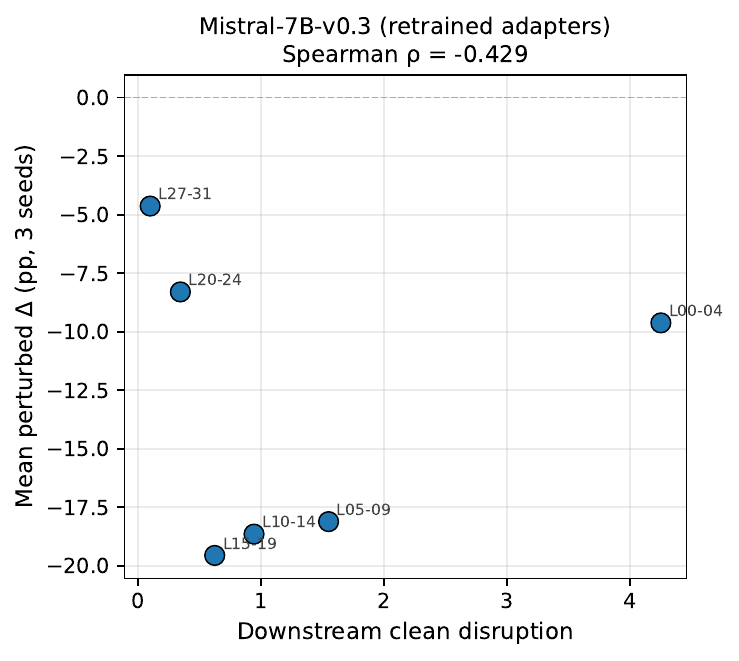}
  \subcaption{Mistral.}
  \label{fig:supp_disruption_scatter_mistral}
\end{subfigure}
\caption{Total clean-input disruption vs.\ mean perturbed $\Delta$ on
independently retrained sweep adapters (both axes from the same
checkpoints; three seeds per window). Direction is consistent in all
four models.}
\label{fig:supp_disruption_scatter}
\end{figure*}

Figure~\ref{fig:supp_disruption_scatter} plots, for the four GSM8K
sweep models (Phi-3.5, Qwen2.5-7B, Llama-3, Mistral), total
clean-input disruption (cf.\ Figure~\ref{fig:disruption} and
Figure~\ref{fig:supp_disruption}) against mean perturbed $\Delta$.
Both axes are measured on a single set of \emph{independently
retrained} sweep adapters (three seeds per window, fixed harness,
\texttt{max\_new\_tokens}${=}768$; the original \S\ref{sec:sweep}
checkpoints were no longer available for disruption measurement).
Higher disruption tracks lower $\Delta$ in all four models (Spearman
$\rho = -0.66 / -0.54 / -0.60 / -0.43$ for Phi/Qwen/Llama/Mistral;
$n{=}6$ windows per model, so no single model reaches significance,
but the direction is consistent in all four).

The retraining doubles as a replication of the layer sweep on fresh
adapters: the qualitative structure of
Table~\ref{tab:layer_sweep_panel} recurs---mid-layer windows worst
(Phi L10--14 $-5.9$, Qwen L08--11 $-7.6$, Llama L10--14 $-12.2$,
Mistral $-18$ to $-20$~pp), deepest windows least harmful---and it
additionally covers Llama-3 and Mistral L10--14, which the original
sweep did not train (both strongly negative, as cascade disruption
predicts). Exact values differ from Table~\ref{tab:layer_sweep_panel}
(fresh training; $768$ vs.\ $512$ generation budget), and the
near-zero late cells fluctuate in sign across retraining (e.g., Qwen
L24--27 $+0.3 \to -0.3$~pp), as expected of effectively null cells.
Gemma-2-9B is omitted (adapter-insensitive; \S\ref{sec:tally}).

\section{Cross-Task LRD Profiles: Phi-3.5 and Mistral}
\label{supp:crosstask_mistral}
\label{sec:crosstask_extra}

\begin{figure*}[!htbp]
\centering
\begin{subfigure}{0.48\textwidth}
  \centering
  \includegraphics[width=\linewidth]{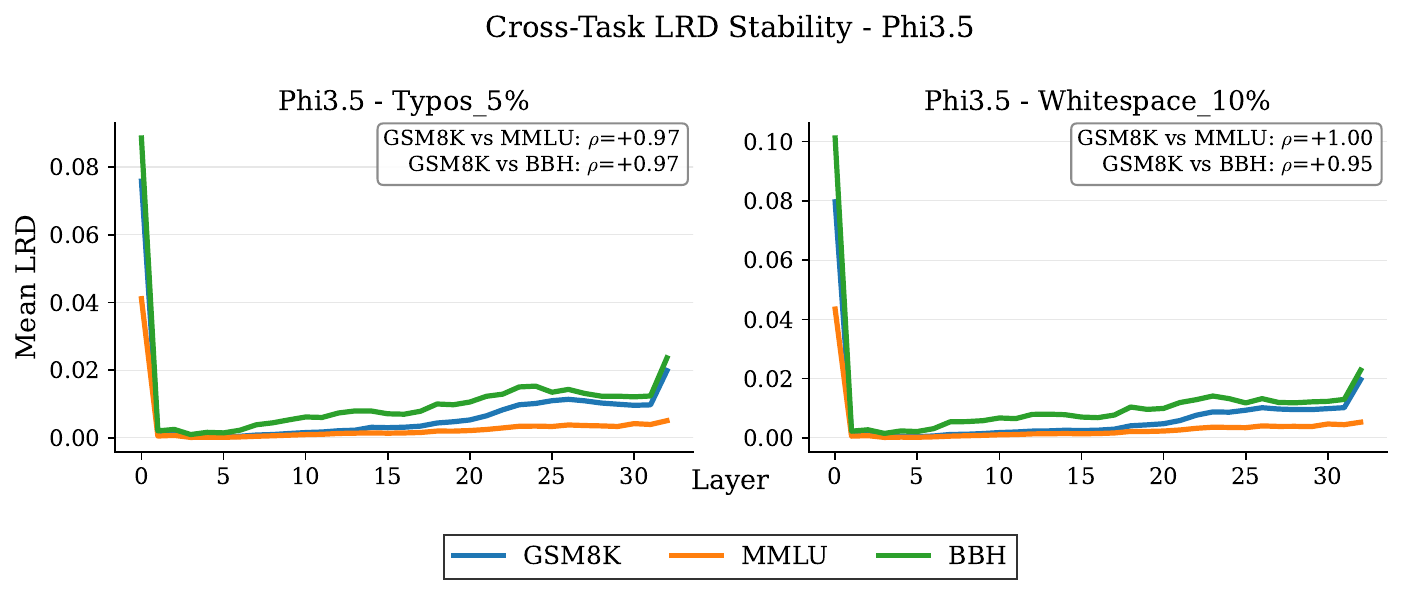}
  \subcaption{Phi-3.5 (spike-and-suppress).}
  \label{fig:supp_crosstask_phi}
\end{subfigure}\hfill
\begin{subfigure}{0.48\textwidth}
  \centering
  \includegraphics[width=\linewidth]{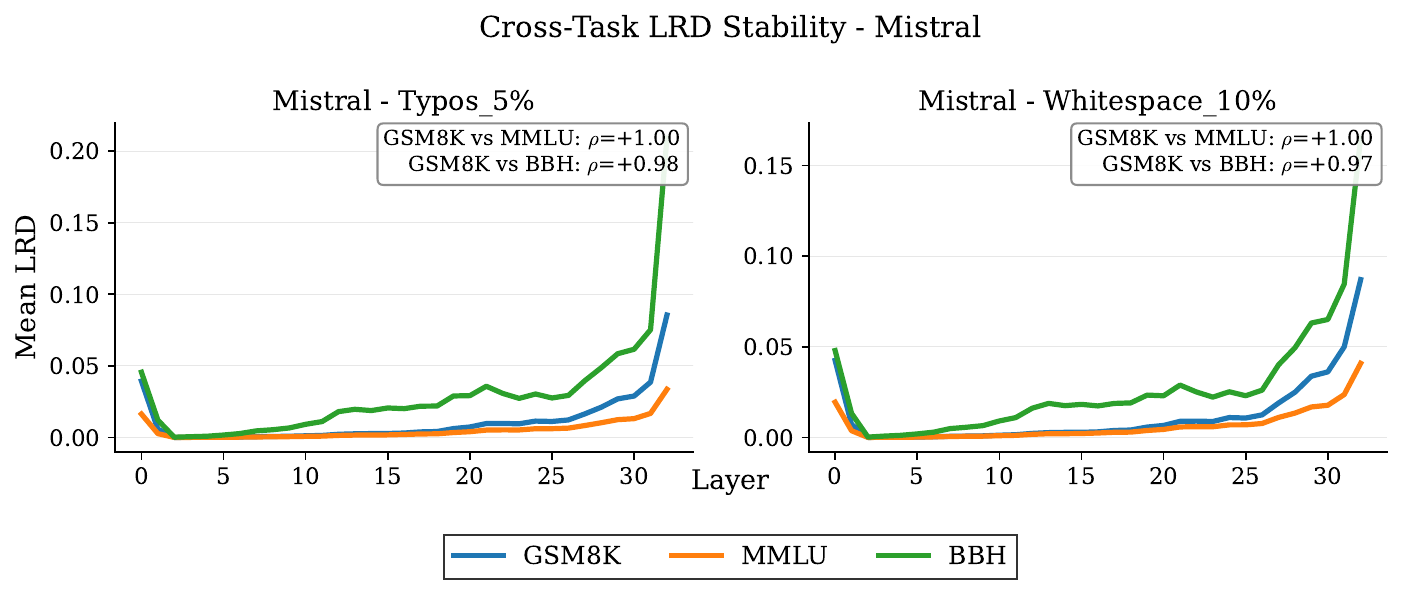}
  \subcaption{Mistral (late-accumulation).}
  \label{fig:supp_crosstask_mistral}
\end{subfigure}
\caption{Cross-task LRD profiles (GSM8K, MMLU, BBH; typos~5\%) for
Phi-3.5 and Mistral. Both produce near-identical shapes across tasks
($\rho > 0.95$).}
\label{fig:supp_crosstask}
\end{figure*}

Figure~\ref{fig:supp_crosstask} shows the cross-task LRD comparison for
the two models on which we collected GSM8K/MMLU/BBH data
(\S\ref{sec:crosstask}). Both models give $\rho > 0.95$ between tasks, indicating that
task modulates LRD magnitude but not profile shape.

\section{Per-Model Layer Sweep Panels}
\label{sec:layer_sweep_per_model}

\begin{figure}[!htbp]
\centering
\includegraphics[width=\linewidth]{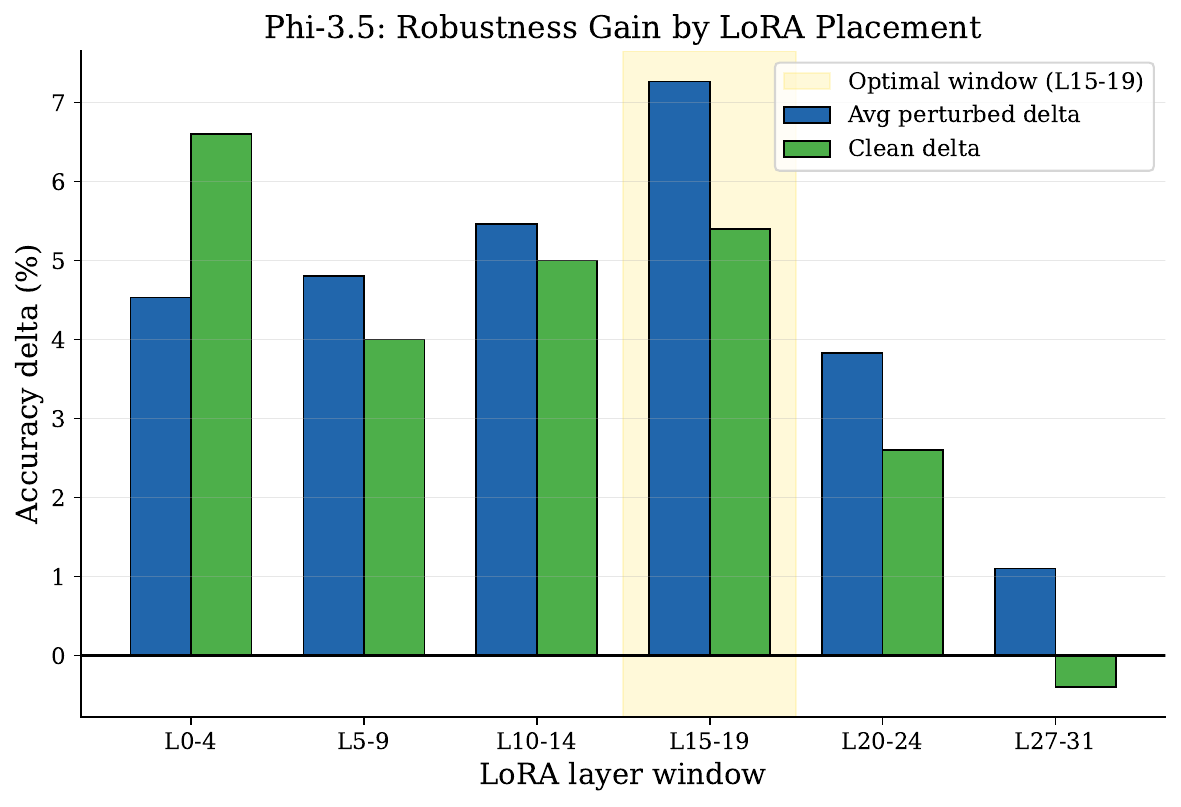}
\caption{Fixed-harness CE-only LoRA layer sweep on GSM8K. Mean
perturbed $\Delta$ (six perturbations, three seeds) on Phi-3.5,
Qwen2.5-7B, Llama-3, and Mistral. On the three adjudicable models the
mid-layer L08--L20 range is uniformly most damaging; non-negative
windows are the deepest available. Mistral's row reflects an lr
artifact (\S\ref{sec:tally}). Gemma-2-9B is adapter-insensitive
($\pm 0.7$~pp) and reported separately.}
\label{fig:layer_sweep}
\end{figure}

\begin{figure*}[!htbp]
\centering
\begin{subfigure}{0.48\textwidth}
  \centering
  \includegraphics[width=\linewidth]{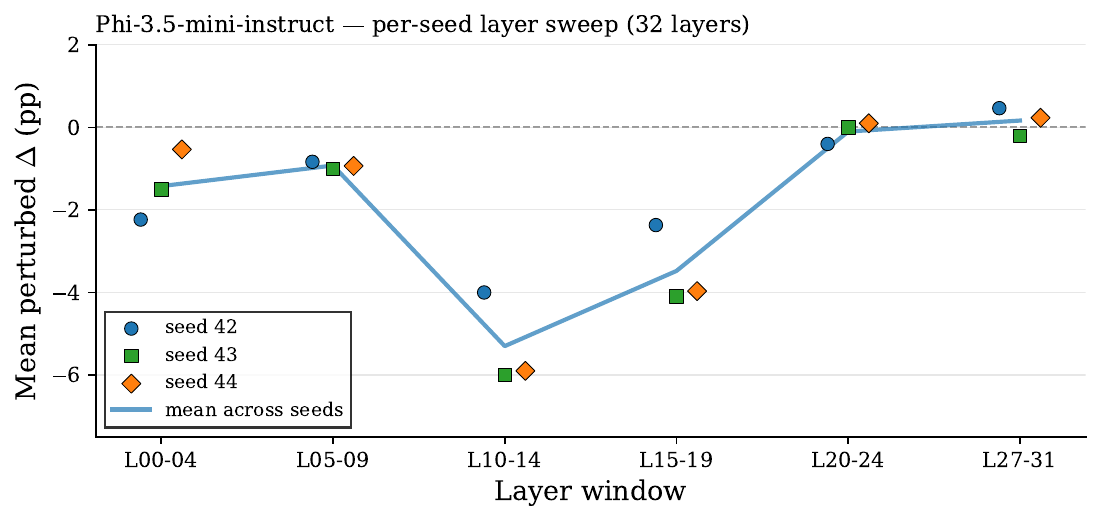}
  \caption{Phi-3.5 (spike-and-suppress; clean $85.4\%$).}
  \label{fig:supp_sweep_phi}
\end{subfigure}\hfill
\begin{subfigure}{0.48\textwidth}
  \centering
  \includegraphics[width=\linewidth]{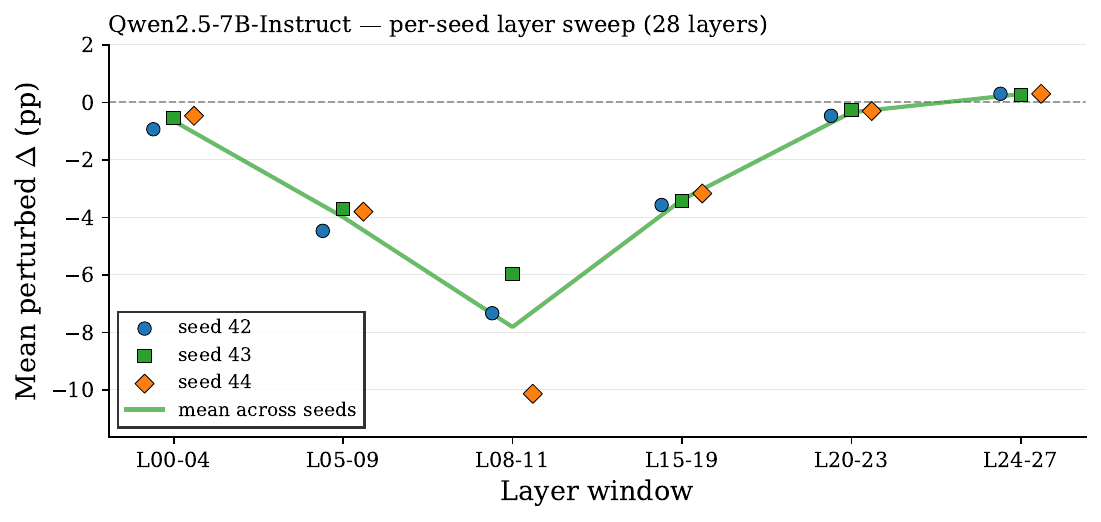}
  \caption{Qwen2.5-7B (late-accumulation; clean $89.0\%$).}
  \label{fig:supp_sweep_qwen}
\end{subfigure}
\\[0.5em]
\begin{subfigure}{0.48\textwidth}
  \centering
  \includegraphics[width=\linewidth]{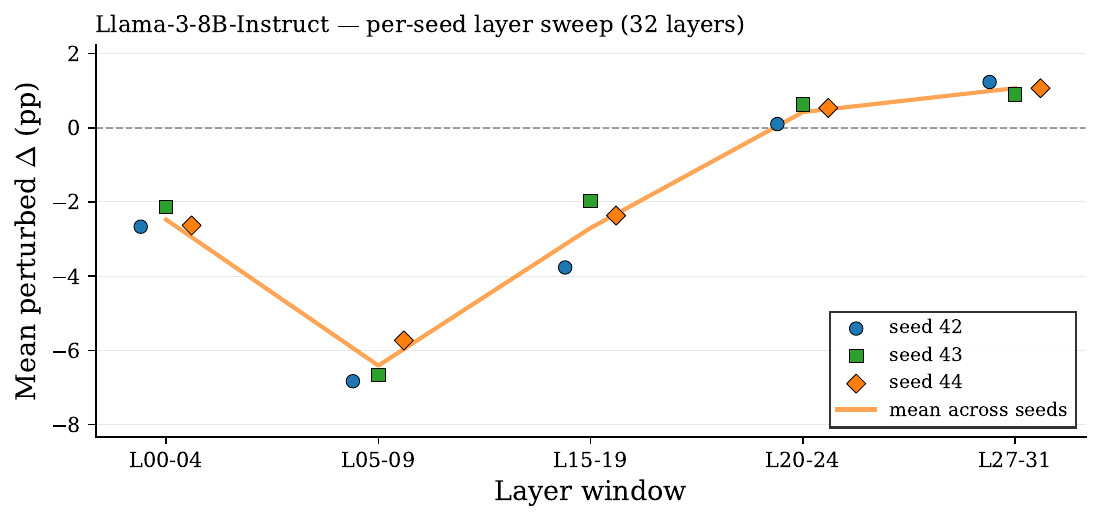}
  \caption{Llama-3-8B (late-accumulation; clean $78.4\%$).}
  \label{fig:supp_sweep_llama}
\end{subfigure}\hfill
\begin{subfigure}{0.48\textwidth}
  \centering
  \includegraphics[width=\linewidth]{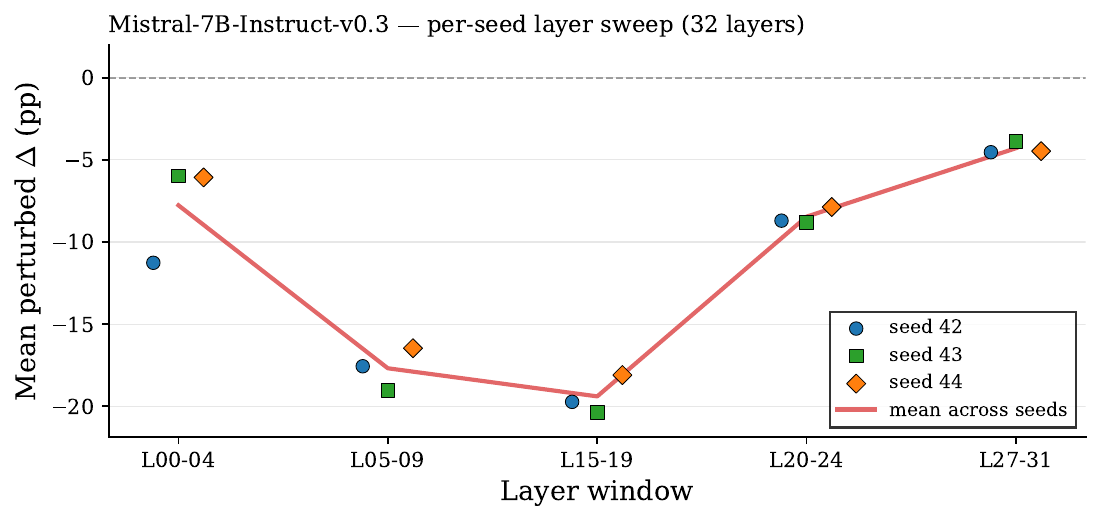}
  \caption{Mistral-7B-v0.3 (late-accumulation; clean $59.6\%$;
  panel outlier).}
  \label{fig:supp_sweep_mistral}
\end{subfigure}
\\[0.5em]
\begin{subfigure}{0.48\textwidth}
  \centering
  \includegraphics[width=\linewidth]{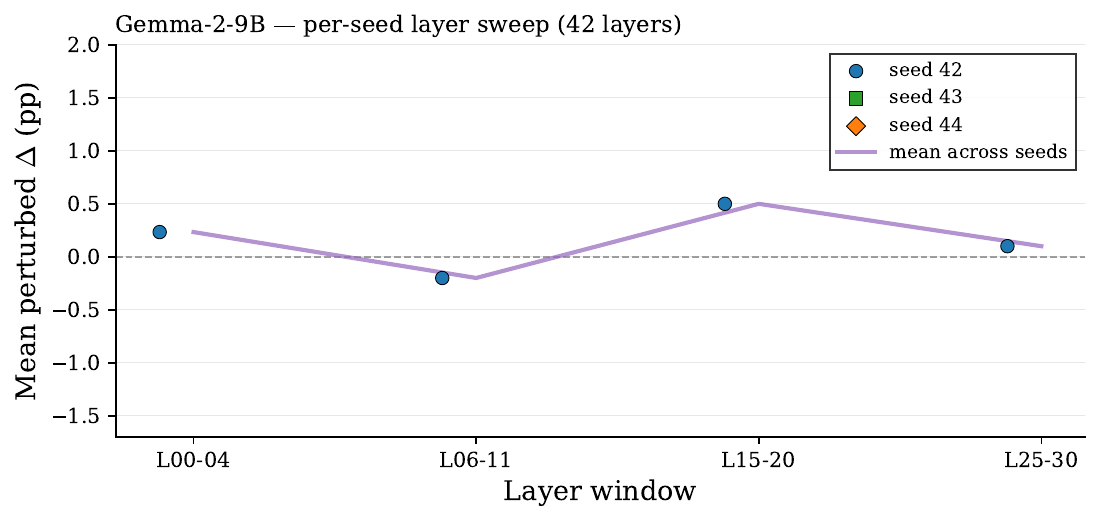}
  \caption{Gemma-2-9B base (spike-and-suppress; clean $64.8\%$;
  single-seed per window with one cross-seed replication at L00--04).}
  \label{fig:supp_sweep_gemma}
\end{subfigure}
\caption{\textbf{Per-Model Layer Sweep Panels.} Fixed-harness CE-only
LoRA, mean perturbed $\Delta$ per non-overlapping $5$-layer window,
three seeds per window (seeds $42$, $43$, $44$). The numerical
summary is in Table~\ref{tab:layer_sweep_panel}.}
\label{fig:supp_layer_sweep_per_model}
\end{figure*}

Figure~\ref{fig:layer_sweep} plots the panel-level sweep summarized
in Table~\ref{tab:layer_sweep_panel};
Figure~\ref{fig:supp_layer_sweep_per_model} shows, for each model in
the panel, the fixed-harness CE-only LoRA layer sweep on GSM8K broken
out by window with three-seed dispersion visible per cell.

 The three adjudicable
models (Phi-3.5, Qwen2.5-7B, Llama-3) share a clear mid-layer wall
(every window in the L05--L20 range is strongly negative), and on
each the deepest available window is the least-regressive. Mistral is
uniformly negative at the shared learning rate (lr artifact;
\S\ref{sec:tally}); Gemma-2-9B is adapter-insensitive across every
window we tested.

Exact windows behind the nominal columns of
Table~\ref{tab:layer_sweep_panel}: Phi-3.5 and Llama-3 use L00--04,
L05--09, L15--19 (Mid), L20--24 (Mid-late), and L27--31 (Late), with
Phi-3.5 adding L10--14 (Early-mid); Llama-3's grid included no
L10--14 window. Qwen2.5-7B uses L00--04, L05--09, L15--19, L20--23,
and L24--27, plus L08--11 (Early-mid)---a pre-registered exploratory
cell at the C3/C4-predicted optimum that overlaps the L05--09
partition cell. Gemma-2-9B uses L00--04, L06--11 (reported under the
nominal L05--09 column), L15--20, L25--30, and L35--40, single-seed
per window (seed-$42$ for L00--04/L06--11/L15--20; seed-$43$ for
L25--30 and L35--40), with L00--04 replicating at seed-$44$
($+0.23$). Dashes in Table~\ref{tab:layer_sweep_panel} mark cells not
trained: Llama-3's grid included no L10--14 window, Gemma's coarser
five-window grid has no early-mid cell, and all-layer baselines ran
only on the three adjudicable models.

\section{Window-Width Ablation, All-Layer Baseline, and MMLU Control}
\label{sec:width_mmlu}

This section reports the three controls referenced in
\S\ref{sec:tally}: the window-width ablation, the all-layer LoRA
baseline, and the MMLU sweep. All use the fixed harness, three seeds
per cell, and the setup of \S\ref{sec:sweep}.

\paragraph{Window-width ablation (Phi-3.5, GSM8K).}
Table~\ref{tab:width_ablation} varies the adapted window width at the
mid and late positions. The sign structure is width-robust---mid
negative at every width, late approximately zero---and mid-window
damage grows monotonically with width.

\begin{table}[H]
\centering
\small
\resizebox{\columnwidth}{!}{\begin{tabular}{lrrr}
\toprule
Window position & W3 & W5 & W7 \\
\midrule
Mid (L10--14 center) & $-2.8 \pm 0.6$ & $-5.3 \pm 1.1$ & $-7.8 \pm 0.4$ \\
Late (L27--31 center) & $-0.0 \pm 0.1$ & $+0.2 \pm 0.3$ & $-0.6 \pm 0.5$ \\
\bottomrule
\end{tabular}}
\caption{Window-width ablation on Phi-3.5 (GSM8K). Mean perturbed
$\Delta$ (pp) $\pm$ std across three seeds; W$n$ = window width $n$.
Width-$3$ windows are
L11--13 (mid) and L28--30 (late); width-$7$ are L09--15 and L25--31
(clamped at the $32$-layer boundary). Width-$5$ values are the
original sweep runs (Table~\ref{tab:layer_sweep_panel},
\texttt{max\_new\_tokens}${=}512$); widths $3$ and $7$ are new runs
under the fixed harness at $768$.}
\label{tab:width_ablation}
\end{table}

\paragraph{All-layer LoRA baseline (GSM8K).}
Table~\ref{tab:all_layer} gives the all-layer baseline underlying the
last column of Table~\ref{tab:layer_sweep_panel}, alongside the
clean-accuracy regression: all-layer adaptation damages clean
performance on all three adjudicable models, as expected if its
early-layer share incurs cascade disruption.

\begin{table}[H]
\centering
\small
\begin{tabular}{lrr}
\toprule
Model & Mean perturbed $\Delta$ & Clean $\Delta$ \\
\midrule
Phi-3.5-mini & $-8.6 \pm 0.7$ & $-8.1 \pm 0.2$ \\
Qwen2.5-7B & $-13.6 \pm 0.1$ & $-14.1 \pm 0.3$ \\
Llama-3-8B & $-9.9 \pm 1.0$ & $-11.4 \pm 0.7$ \\
\bottomrule
\end{tabular}
\caption{All-layer LoRA baseline (GSM8K, fixed harness). Mean
$\Delta$ (pp) $\pm$ std across three seeds, relative to the
no-adapter baseline. Qwen's tight std reflects distinct checkpoints
whose six-condition means nearly coincide (per-condition accuracies
differ by up to $3$~pp across seeds); all-layer adaptation leaves no
seed-dependent window choice to vary.}
\label{tab:all_layer}
\end{table}

\paragraph{MMLU control sweep.}
Table~\ref{tab:mmlu_sweep} reports the fixed-harness sweep on MMLU
for one model per regime. Effects are near-zero throughout; the only
reliably nonzero cell is Phi-3.5's diagnostic-flagged mid window
L10--14 ($-1.8 \pm 0.4$~pp). No window produces positive transfer on
either model. See \S\ref{sec:tally} for interpretation; we do not
claim the GSM8K mid-layer wall replicates on MMLU.

\begin{table}[H]
\centering
\small
\begin{tabular}{lr@{\hspace{2em}}lr}
\toprule
\multicolumn{2}{c}{Phi-3.5-mini} & \multicolumn{2}{c}{Qwen2.5-7B} \\
Window & $\Delta$ (MMLU) & Window & $\Delta$ (MMLU) \\
\midrule
L00--04 & $+0.4 \pm 0.2$ & L00--04 & $+0.0 \pm 0.1$ \\
L05--09 & $+0.4 \pm 0.1$ & L05--09 & $+0.1 \pm 0.0$ \\
L10--14 & $-1.8 \pm 0.4$ & L08--11 & $-0.0 \pm 0.1$ \\
L15--19 & $-0.4 \pm 0.3$ & L15--19 & $+0.0 \pm 0.1$ \\
L20--24 & $-0.3 \pm 0.1$ & L20--23 & $+0.1 \pm 0.1$ \\
L27--31 & $-0.0 \pm 0.1$ & L24--27 & $-0.1 \pm 0.1$ \\
\bottomrule
\end{tabular}
\caption{MMLU layer sweep (fixed harness, one model per regime). Mean
perturbed $\Delta$ (pp) $\pm$ std across three seeds per window.}
\label{tab:mmlu_sweep}
\end{table}

\section{Earlier Investigations (Pre-Fixed-Harness)}
\label{sec:earlier_investigations}
\label{supp:earlier}
\label{supp:lora_capacity}
\label{sec:capacity}
\label{supp:loss_geometry}
\label{supp:matched_rate}
\label{supp:transfer}
\label{supp:lstab}
\label{sec:lambda}
\label{supp:alt_stabilizer}
\label{sec:alt_stabilizer}
\label{supp:versions}
\label{sec:version}

\begin{figure}[!htbp]
\centering
\includegraphics[width=\linewidth]{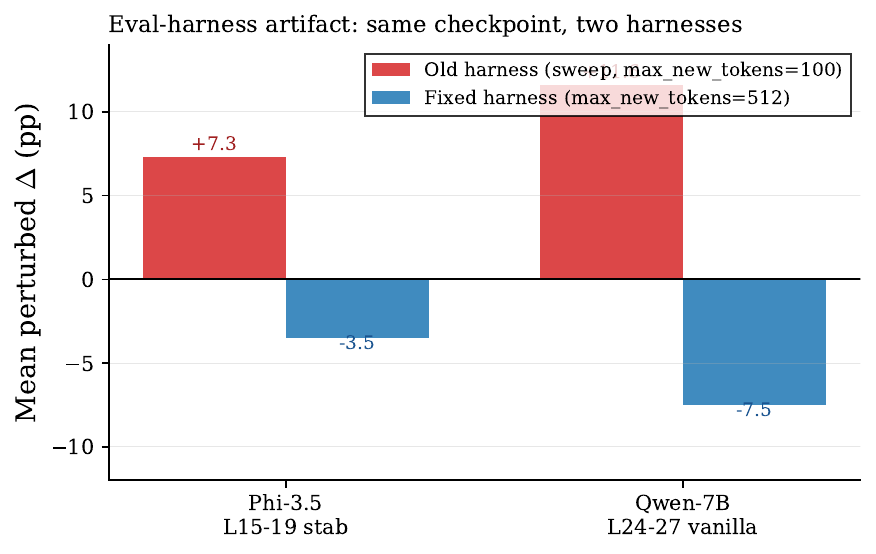}
\caption{Same checkpoints, two harnesses (\S\ref{sec:negative}). Mean
perturbed $\Delta$ for Phi-3.5 L15--19 with the stability loss
(earlier harness: $+7.3$~pp) and the Qwen2.5-7B L24--27 vanilla CE
baseline (earlier harness: $+11.6$~pp), under the earlier
\texttt{max\_new\_tokens}${=}100$ harness (red) vs.\ the fixed harness
(blue); $n{=}500$. The fixed harness uses
\texttt{max\_new\_tokens}${=}512$ for this checkpoint comparison and
the layer sweep (Table~\ref{tab:layer_sweep_panel}); per-experiment
settings in Table~\ref{tab:supp_repro}. The earlier gains were truncated
chain-of-thought scored as empty.}
\label{fig:harness_artifact}
\end{figure}

This section catalogs investigations carried out before the evaluation
harness bug was identified (\S\ref{sec:negative}). All numerical results in
those experiments were generated by an evaluation harness with
\texttt{max\_new\_tokens=100}, which truncated chain-of-thought
generations before the \texttt{\#\#\#\# NUMBER} extraction line and
scored truncated answers as empty. The same checkpoints read between
$+10$ and $+20$~pp differently under the fixed harness, depending on
condition (see Figure~\ref{fig:harness_artifact} and
\S\ref{sec:negative}). The qualitative
conclusions of these investigations are preserved below; full
numerical tables are omitted from this revision because they are not
trustworthy on absolute scale.

\paragraph{Cosine versus MSE stability loss, with matched-rate control.}
We ran cosine and MSE variants of the representation-stability loss at
the original variable per-perturbation rates and at a matched $10\%$
rate. At the original rates, a per-condition split appeared (cosine
winning on typos/OCR/speech/homophones, MSE on whitespace/case),
which we had provisionally interpreted as a taxonomy: ``directional''
perturbation types suited to a cosine loss versus ``uniform'' types
suited to MSE. The matched-rate control collapsed the split: Cohen's
$d$ values for the two groups overlapped completely ($0.40$--$0.48$
vs.\ $0.15$--$0.48$), and the pooled interaction CI went from
$[+2.5\%, +8.7\%]$ at variable rates to $[-2.5\%, +3.5\%]$ at matched
rates. The taxonomy claim from this loss-geometry work was therefore
disproved by the matched-rate control; the underlying intervention has
now also been disproved by the harness fix.

\paragraph{$\lambda_{\text{stab}}$ scaling sanity check.}
At $\lambda_{\text{stab}} = 0.3$, $L_{\text{stab}}$ contributed
$\leq 0.6\%$ of the total loss in every run. Setting
$\lambda_{\text{stab}} = 3.0$ raised the contribution to $6$--$8\%$.
This established that earlier cosine-vs-MSE comparisons at
$\lambda_{\text{stab}} = 0.3$ were essentially noise. Under the
fixed-harness ablation reported in the main paper, however, even the
properly weighted $\lambda_{\text{stab}} = 1.0$ regime \emph{hurts} CE
LoRA by $1.2$--$3.9$~pp on the conditions and seeds we tested. The
stability loss is not a useful component once the evaluation harness is
correct.

\paragraph{LoRA capacity experiments.}
We swept LoRA ranks $\{16, 32, 64\}$ at $\{3{,}000, 5{,}000\}$ training
steps on Llama-3 and Mistral and observed monotonic degradation with
rank and steps. The qualitative conclusion---that the prior stabilizer's
failure on late-accumulation models is not a capacity issue---survives
the harness fix. A separate ``capacity concentration'' result claimed
that a concentrated $5$-layer LoRA at L15--L19 outperforms an all-layer
LoRA by $+5.1$~pp perturbed; that $+5.1$ number is broken-harness and
is excluded from this revision. The cascade-disruption explanation in
the main paper does not depend on it.

\paragraph{Transfer evaluation.}
GSM8K-trained adapters with the prior stability loss degraded MMLU by
$1.5$--$4.5$~pp and were near-zero on BBH. Task-specific MMLU training
produced near-zero average gains. Diagnostic-side cross-task stability
(LRD profile $\rho > 0.95$ across GSM8K/MMLU/BBH on Phi-3.5 and Mistral)
is harness-independent and is retained in the main paper.

\paragraph{Alternative residual-adapter stabilizer.}
Independently of the LoRA-based stabilizer, we tested a lightweight
two-layer bottleneck residual adapter $x \mapsto x + g \cdot W_2
\sigma(W_1 x)$ inserted after the attention block at four placement
configurations on Qwen-7B (layers 2/6/10/14; 8/12/16/20; 12/16/20/24;
4/10/18/26) and at the first eight layers on Mistral. Training mixed
clean and OCR-perturbed inputs with clean targets. Every configuration
regressed: Qwen-7B baseline outperformed all four stabilized variants
by approximately $9$--$11$~pp on GSM8K OCR-$5\%$; Mistral baseline
exceeded the stabilized variant by approximately $4.5$~pp. The
qualitative result (multiple stabilizer designs, including a non-LoRA
one, fail to improve robustness on these models) is consistent with the
main-paper negative result. As with the rest of this section, the
absolute numbers are pre-fixed-harness.

Full per-condition numbers from each of these investigations are
available on request and will be released with the public code drop.

\section{Reproducibility Settings}
\label{supp:repro}
\label{sec:repro}

\begin{table}[H]
\centering
\small
\resizebox{\columnwidth}{!}{\begin{tabular}{ll}
\toprule
Component & Configuration \\
\midrule
Eval harness                         & CoT, \texttt{\#\#\#\#} extraction \\
\texttt{max\_new\_tokens}            & $512$: layer sweep (Table~\ref{tab:layer_sweep_panel}), two-harness comparison \\
                                     & $768$: diagnostics, scaling, width-$3/7$, all-layer \\
                                     & $32$: MMLU \\
Diagnostics sample size              & $n=500$ per condition \\
Scale-variation sample size          & $n=500$ test; $n_\text{clean-correct}{=}200$ \\
Patching pairs                       & Phi-3.5 $n{=}50$; Llama-3 $n{=}100$; Qwen-14B $n{=}84/100$ \\
LoRA rank / alpha                    & $4$ / $8$ \\
LoRA targets                         & \texttt{q\_proj}, \texttt{v\_proj} \\
LoRA loss                            & cross-entropy only ($\lambda_{\text{stab}}{=}0$) \\
Layer-sweep windows                  & non-overlapping $5$-layer windows across depth \\
Layer-sweep seeds                    & $\{42, 43, 44\}$ (three per window) \\
Training steps                       & $300$ (Phi-3.5); $1{,}000$--$3{,}000$ (Llama-3, Mistral, Qwen, Gemma) \\
Bootstrap                            & block size~$5$, $10{,}000$ resamples \\
\midrule
Hardware                             & NVIDIA A40 ($46$GB), A100 ($80$GB), V100 ($32$GB); single-GPU jobs \\
Cluster                              & SLURM \\
Total compute                        & ${\approx}900$ GPU-hours (reported experiments; excludes exploratory and failed runs) \\
\bottomrule
\end{tabular}}
\caption{Reproducibility settings for the fixed-harness experiments in
the main paper. The $\lambda_{\text{stab}}{=}1.0$ ablation referenced
in \S\ref{sec:negative} uses the same settings with the stability-loss
term re-enabled. The pre-fixed-harness investigations in
\S\ref{sec:earlier_investigations} used different settings
(\texttt{max\_new\_tokens}${=}100$, $\lambda_{\text{stab}}{=}3.0$, fixed
optimal windows).}
\label{tab:supp_repro}
\end{table}

\paragraph{MMLU control sweep.}
The MMLU sweep (\S\ref{sec:tally}, Table~\ref{tab:mmlu_sweep}) uses
\texttt{max\_new\_tokens}${=}32$ and a fixed evaluation subset; the
item IDs for that subset are included in the released code.

\paragraph{Software stack.}
Single-GPU PyTorch~\texttt{2.10.0+cu128} with the following libraries:
\texttt{transformers}~\texttt{4.44.2},
\texttt{peft}~\texttt{0.11.0},
\texttt{datasets}~\texttt{4.5.0},
\texttt{accelerate}~\texttt{1.12.0},
\texttt{numpy}~\texttt{2.2.5},
\texttt{scipy}~\texttt{1.15.3}. Exact pinned versions and the full
\texttt{requirements.txt} accompany the code release.

\paragraph{Model checkpoints and loading.}
Weights are loaded from the official HuggingFace releases:
\texttt{microsoft/Phi-3.5-mini-instruct},
\texttt{mistralai/Mistral-7B-Instruct-v0.3},
\texttt{meta-llama/Meta-Llama-3-8B-Instruct},
\texttt{google/gemma-2-9b},
\texttt{Qwen/Qwen2.5-7B-Instruct},
\texttt{Qwen/Qwen2.5-1.5B-Instruct},
\texttt{Qwen/Qwen2.5-14B-Instruct}, and
\texttt{meta-llama/Llama-3.2-1B-Instruct}. All models load in
\texttt{torch\_dtype=bfloat16} with \texttt{attn\_implementation="eager"}
where required by the activation-patching hooks.

\paragraph{LoRA configuration (\texttt{peft}).}
\texttt{r=4}, \texttt{lora\_alpha=8},
\texttt{target\_modules=["q\_proj","v\_proj"]} for Mistral, Llama-3,
Gemma-2, and all Qwen scales;
\texttt{target\_modules=["qkv\_proj","o\_proj"]} for Phi-3.5-mini
(fused attention). Inference-mode flag toggled per call; otherwise
default \texttt{peft} arguments.

\paragraph{Datasets (\texttt{datasets}).}
\texttt{openai/gsm8k} (main config, test split) drawn with shuffle
seed~$42$ and \texttt{n\_samples}~$\in\{200,500\}$ per condition as
reported in Table~\ref{tab:supp_repro}; MMLU and BBH loaded from
their respective official HuggingFace releases (cited in
\S\ref{sec:diagnostics}).

\paragraph{Optimizer (\texttt{torch}).}
\texttt{torch.optim.AdamW} with default $(\beta_1,\beta_2,\epsilon)$
and weight decay; learning rates as specified in the main paper
(shared-grid sweep at $5\mathrm{e}{-}5$, Mistral audit at
$1\mathrm{e}{-}5$). Training in \texttt{bfloat16} except where noted.

\paragraph{Statistics (\texttt{scipy}, \texttt{numpy}).}
\texttt{scipy.stats.pointbiserialr},
\texttt{scipy.stats.mannwhitneyu},
\texttt{scipy.stats.levene}, and \texttt{scipy.stats.spearmanr} are
used at default arguments for the per-perturbation significance
checks and the pairwise correlations reported above. The
moving-block bootstrap is an in-house NumPy implementation (block
size~$5$, $10$k resamples).

\end{document}